\documentclass[10pt,journal,compsoc]{IEEEtran}
\usepackage{amsmath,amsfonts}
\usepackage{algorithmic}
\usepackage{algorithm}
\usepackage{array}
\usepackage[caption=false,font=normalsize,labelfont=sf,textfont=sf]{subfig}
\usepackage{textcomp}
\usepackage{stfloats}
\usepackage{url}
\usepackage{verbatim}
\usepackage{graphicx}
\usepackage{cite}

\usepackage{graphicx}

\usepackage{bm}
\usepackage{multirow}
\usepackage{booktabs} 
\usepackage{graphicx} 
\usepackage{caption} 
\usepackage{amsmath}
\usepackage{amssymb}
\usepackage{bm}
\usepackage{array}
\usepackage{tabularx}
\usepackage{booktabs}
\usepackage{multirow}
\usepackage{adjustbox}
\usepackage{orcidlink}
\usepackage{xcolor}
\usepackage{graphicx}
\usepackage{pifont}
\newcommand{\cmark}{\ding{51}}
\newcommand{\xmark}{\ding{55}}
\begin{document}

\title{FunHOI: Annotation-Free 3D Hand-Object Interaction Generation via Functional
Text Guidance}

\author{Yongqi~Tian$^{\dagger}$,
        Xueyu~Sun$^{\dagger}$,
        Haoyuan~He,
        Jianlei~Wang,
        and~Caigui~Jiang$^{*}$%
\thanks{Yongqi Tian, Xueyu Sun, Haoyuan He, Jianlei Wang, and Caigui Jiang are with the State Key Laboratory of Human-Machine Hybrid Augmented Intelligence, Institute of Artificial Intelligence and Robotics, Xi'an Jiaotong University, Xi'an, China.}
\thanks{$^{\dagger}$Yongqi Tian and Xueyu Sun contributed equally to this work.}
\thanks{$^{*}$Corresponding author: Caigui Jiang.}
\thanks{E-mail: yongqitian@stu.xjtu.edu.cn, xueyusun@stu.xjtu.edu.cn, hehy02@stu.xjtu.edu.cn, Wjl\_191013@stu.xjtu.edu.cn, cgjiang@xjtu.edu.cn.}
}

\markboth{IEEE Transactions on Visualization and Computer Graphics}%
{Tian \MakeLowercase{\textit{et al.}}: FunHOI: Annotation-Free 3D Hand-Object Interaction Generation via Functional Text Guidance}

\maketitle
\begin{abstract}
Hand-object interaction (HOI) is central to human manipulation, but synthesizing functional 3D HOI remains challenging due to dexterous hand motion, complex contact patterns, and strong geometric ambiguity. Existing methods often rely on 3D HOI annotations, predefined object templates, or supervised interaction priors. Such requirements limit their applicability to open-ended functional grasping scenarios, where the same object may require different contact regions and poses depending on the intended use. We present FunHOI, an annotation-free two-stage framework for functional-intent-conditioned 3D HOI synthesis from natural-language prompts. FunHOI represents each prompt using action--object--purpose semantics and first translates the functional intent into a semantically aligned 2D HOI guidance image through the Functional Grasp Generator (FGG). Rather than learning a direct text-to-3D generative prior, FunHOI reconstructs the hand and object from this 2D semantic guidance and composes them in 3D using the Functional Grasp Refiner (FGR), which performs scale-aware alignment, object pose approximation, and contact-driven refinement to improve spatial consistency and physical plausibility. Extensive experiments show that FunHOI achieves competitive reconstruction quality, improved hand-object contact plausibility, and better alignment with functional intent, without requiring 3D HOI annotations.
\end{abstract}

\IEEEdisplaynontitleabstractindextext
\IEEEpeerreviewmaketitle

\begin{figure*}[!t]
\centering
\includegraphics[width=\textwidth]{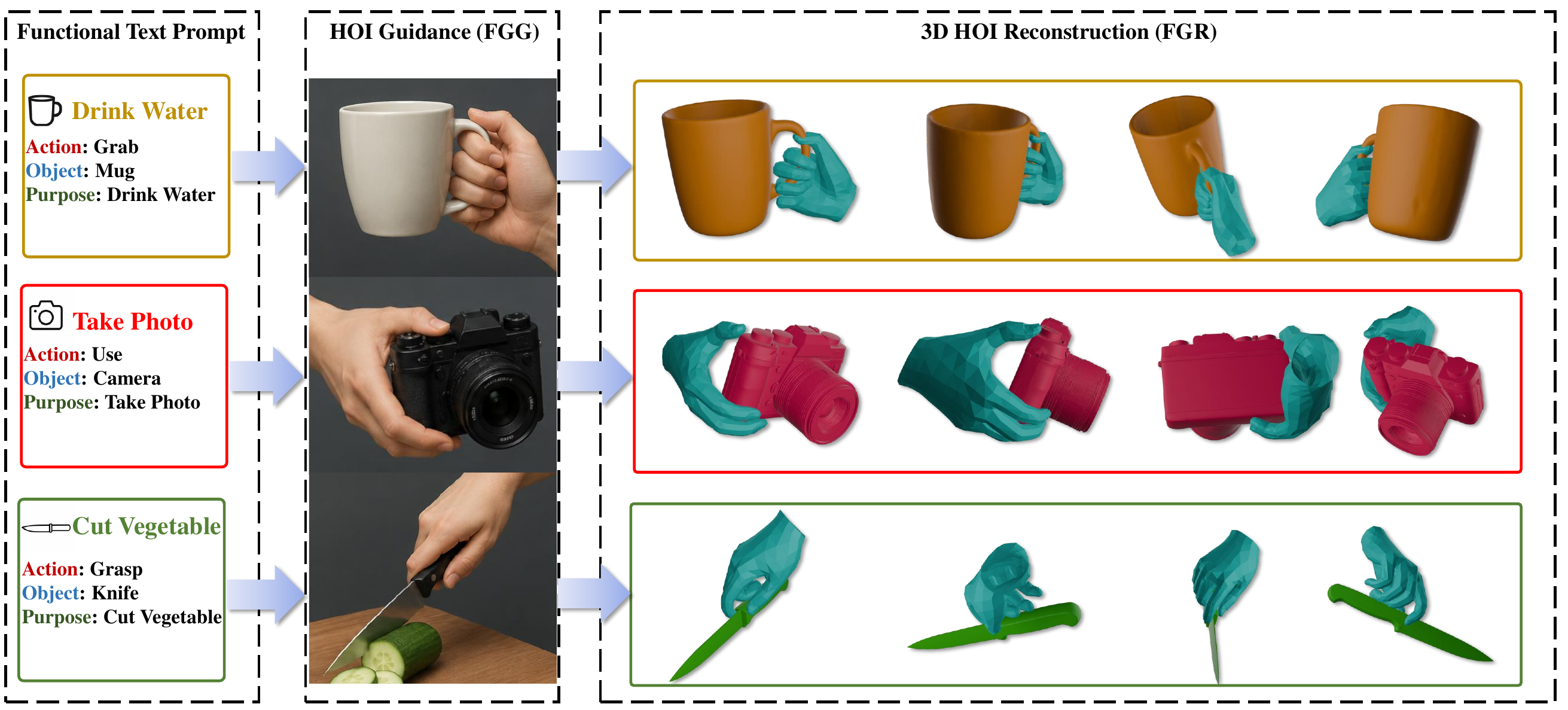}
\caption{Representative results of FunHOI from functional text prompts to 3D hand-object interactions. Given a functional prompt, FGG generates a semantically aligned 2D HOI guidance image, and FGR reconstructs and refines the corresponding 3D hand-object interaction from multiple viewpoints. The examples show functional interactions for drinking from a mug, taking a photo with a camera, and cutting vegetables with a knife.}
\label{fig:teaser}
\end{figure*}

\section{Introduction}

\IEEEPARstart{S}{erving} as the core medium for human--environment interaction, the hand executes multi-dimensional tasks ranging from basic survival acts (e.g., \textit{drinking water}) to complex tool operations (e.g., \textit{taking photos}). 
These interactions require dexterous hand motion, appropriate object pose, and physically plausible contact that is consistent with the intended object use.
Addressing the key challenge of interactive semantic parsing in Embodied Intelligence and Artificial Intelligence Generated Content research, this study proposes a natural language-driven framework for functional 3D hand--object interaction (HOI) generation. 
By decoupling intention-embedded semantic descriptions from explicit geometric supervision, our framework enables functional 3D HOI reconstruction without requiring any 3D HOI annotations (see Fig.~\ref{fig:teaser} for sample results).

The 3D human--object interaction (HOI) generation task aims to produce physically plausible interactions that reflect human intentionality while maintaining scene functionality. 
This task involves two fundamental challenges. 
The first is \textbf{how to grasp}, which concerns the configuration of the hand and is typically conditioned on both the task description and object-specific characteristics. 
The task description provides high-level contextual cues for the intended interaction, while object properties constrain feasible hand shapes. 
The second is \textbf{where to grasp}, which focuses on localizing appropriate contact regions on the object surface and depends jointly on the functional intent expressed in the text and the geometric structure of the object to ensure effective and accurate grasping. 
Previous approaches~\cite{text2grasp,diffhoi,song2023learning} have explored the relationship between text and grasping behaviors. 
However, most existing methods rely on explicit 3D annotations and positional commands (e.g., ``grasp the handle of the mug''), which encode grasping as predefined mechanical actions rather than intent-driven interactions (e.g., ``grasp the mug to drink''). 
For the same object, different purposes may require different contact regions and object poses, such as drinking from a mug versus pouring from it.
As a result, these methods are constrained to fixed input templates and limited object categories, making it difficult to generalize to diverse functional intents and open-ended interaction scenarios.

In addition, obtaining relevant data remains a significant challenge in this task. 
In recent years, an increasing number of HOI datasets~\cite{hoi4d,ho3D1,ho3D2,hasson19_obman,ycb,damen2018scaling,brahmbhatt2020contactpose,corona2020ganhand,hasson2019learning,jian2023affordpose,taheri2020grab,yang2022oakink} have been created to facilitate machine learning research on human grasping behaviors~\cite{jian2023affordpose,hasson2021towards,yang2022oakink}. 
Some studies~\cite{choi,gsdf,alignsdf,10093013} attempted to train models to understand object interactions using publicly available 3D annotated data. 
However, these datasets often require complex motion capture devices and substantial resource investment. 
Previous methods~\cite{hasson2021towards,hold,diffhoi,brahmbhatt2019contactdb} managed to achieve high-quality 3D HOI reconstruction using video data. 
However, due to the lack of fine-grained functional annotations in the training data, existing methods struggle to establish robust semantic association models and heavily rely on predefined object category spaces. 
This limitation significantly hinders the generalization capability of the algorithms, particularly in open-ended scenarios, where performance degradation is most prominent when handling single images generated by text-to-image models. 
Specifically, the current HOI datasets fail to comprehensively capture the diversity of human behaviors, which makes learning-based HOI reconstruction algorithms struggle to adapt to increasingly complex hand-object interaction scenarios. 
As a result, performing functional 3D HOI reconstruction solely from a single 2D image remains an exceptionally challenging task.

We propose \textbf{FunHOI}, a two-stage, annotation-free framework for functional-intent-conditioned 3D hand--object interaction (HOI) synthesis. 
Rather than introducing a new network architecture, FunHOI focuses on translating functional intent expressed in natural language into physically plausible 3D interactions, without requiring 3D HOI annotations. 
We clarify that our framework performs text-conditioned 3D reconstruction via 2D semantic guidance, rather than learning a direct generative 3D prior from text.

Our framework consists of two stages. 
The first stage, termed the \textbf{Functional Grasp Generator (FGG)}, maps functional text descriptions to a 2D HOI guidance image together with initial 3D reconstructions of the hand and object. 
The second stage, the \textbf{Functional Grasp Refiner (FGR)}, composes the hand and object into a coherent 3D interaction by enforcing scale consistency, relative pose alignment, and contact plausibility through a structured optimization process.
Instead of directly performing 3D reconstruction from images, we leverage text-to-image generation guided by functional descriptions as an intermediate representation, effectively bridging high-level semantics and 3D interaction composition.

The main contributions of this work are as follows:
\begin{itemize}
  \item We introduce \emph{functional intent}, explicitly decomposed into action--object--purpose, as a high-level control signal for 3D HOI generation, enabling intent-driven interactions beyond template-based or position-based commands.
  \item We propose FunHOI, a two-stage, annotation-free framework that translates functional text into physically plausible 3D hand--object interactions without requiring 3D HOI annotations.
  \item We design a structure-aware alignment and refinement mechanism that jointly addresses scale consistency, relative pose estimation, and hand--object contact plausibility, bridging the gap between 2D semantic HOI guidance and 3D interaction composition.
\end{itemize}

\section{Related Work}

3D HOI Reconstruction aims to construct three-dimensional scenes that encompass the complete geometry of the hand and the object, as well as their patial pose that implies the semantics of interaction. 

\textbf{Template-based 3D HOI Reconstruction} focuses on the relative spatial relationship between the hand and the object. These methods adjust the pose of the hand given a predefined object geometry model to meet interaction demands~\cite{text2grasp,dexgraspnet}. 
While \textbf{Template-free 3D HOI Reconstruction} treats the construction of the object's geometry as an integral part of the reconstruction task. Such approaches generate the object's geometry conditioned on various inputs, e.g., text or images, and subsequently construct the complete 3D HOI scene on this basis~\cite{gsdf,liu2024easyhoi,ihoi,zhang2024moho}. 
Recent advances such as MagicHOI~\cite{magichoi2025} and Hand-held Object Reconstruction with Dynamic Interaction~\cite{jiang2025dynhor} leverage novel view synthesis priors and dynamic interaction cues to regularize occluded regions and improve robust 3D hand-object geometry and pose reconstruction from monocular interaction videos under fast motion and severe occlusions.

\textbf{Physical 3D HOI Reconstruction.} Several studies on 3D HOI reconstruction have focused on integrating physical and spatial guidance to ensure the production of physically plausible scenarios~\cite{dexgraspnet, grady2021contactopt, weng2024dexdiffuser,wang2013video}. 
Recent force-aware reconstruction frameworks such as ViTaM-D~\cite{yu2025vitamd} introduce distributed contact and force representations to refine interaction details and improve physical realism, particularly in complex dynamic and occluded contact regions.
These approaches optimize the physical interactions between hands and objects, aiming to generate 3D HOI data that adhere to established physical laws for enhanced realism. Inspired by these efforts, our work further refines the reconstruction process by introducing a set of combined loss terms specifically designed to enhance the realism of 3D HOI.

\textbf{Task-Oriented 3D HOI Reconstruction.} Task-oriented 3D HOI Reconstruction aims to model the connection between {abstract high-level task semantics} and dexterous hand-object interactions, targeting different tasks for the same category of objects. Recent works combine 3D segmentation frameworks to provide more fine-grained task-semantic-related optimization objectives, synthesizing HOI scenes for different high-level abstract tasks~\cite{text2grasp, wei2024grasp, chen2023task, wei2023generalized, zhu2021toward,sohn2015learning,karunratanakul2020grasping}. However, these methods intuitively introduce additional data annotation, which results in a labor-intensive workload. Our method, without requiring 3D HOI annotations, extends 3D HOI reconstruction to generate task-oriented hand--object interactions from different abstract task descriptions.

Overall, FunHOI targets functional-intent-conditioned 3D HOI synthesis from action--object--purpose prompts, without relying on task-specific 3D HOI annotations or predefined object templates.

\section{Method}

\begin{figure*}[htbp]
  \centering
  \includegraphics[width=\textwidth]{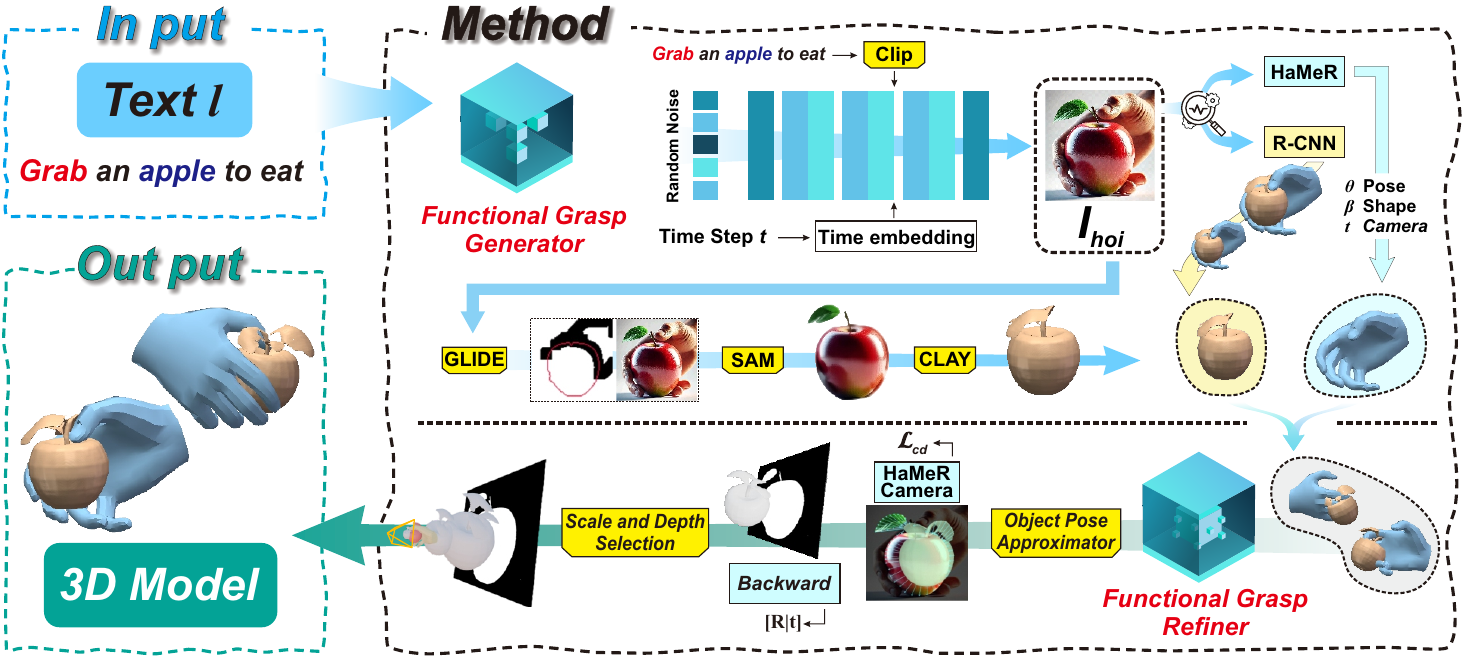} 
  \caption{
  Overview of FunHOI. Given a functional text prompt, FGG generates a semantically aligned 2D HOI guidance image and obtains initial hand/object reconstructions. FGR then composes and refines the 3D hand--object interaction through scale alignment, object pose approximation, and contact-aware optimization.
  }
  \label{fig:2}
\end{figure*}

Given a functional text $l$, our framework aims to generate a corresponding 3D HOI model. To this end, we propose a two-stage pipeline, \textbf{Functional HOI (FunHOI)}. In the first stage, the \textbf{Functional Grasp Generator (FGG)} generates a 2D HOI guidance image, corresponding hand-object meshes, camera parameters and the hand parameters represented by MANO\cite{MANO:SIGGRAPHASIA:2017}. In the second stage, the \textbf{Functional Grasp Refiner (FGR)} further composes the hand and object meshes into a 3D HOI scene consistent with the generated 2D HOI guidance while ensuring physical plausibility. An overview of the proposed framework is illustrated in Fig.~\ref{fig:2}.

The subsequent sections are organized as follows: In Sec.\ref{sec:p}, we present the fine-tuning strategy of the diffusion model and the method for generating functional text prompts. We explicitly introduce the geometric reconstruction of hand-object in Sec.~\ref{sec:3.1}. Finally, in Section~\ref{sec:3.2}, we provide an in-depth explanation of the FGR refining process for HOI we developed.

\subsection{Preliminary Work}\label{sec:p}
Considering the complexity of interaction scenarios, viewpoint discrepancies, and the high cost of data annotation, acquiring 3D prior information for HOI poses significant challenges. This study adopts a diffusion model~\cite{rombach2022high} to learn the relationship between functional text and HOI images. Specifically, the model generates the 2D HOI image $I_{hoi}$ from the input functional text, and infers 3D HOI configurations based on the generated 2D images. To achieve this, we need to prepare HOI images and their corresponding functional text pairs for fine-tuning the diffusion model.

To enable the diffusion model to better capture the characteristics of HOI scenes while retaining its generalization ability, we employ DreamBooth~\cite{ruiz2023DreamBooth} for fine-tuning. DreamBooth is a technique for customizing a pre-trained text-to-image model using a small set of reference images. 
In our setting, the reference images correspond to functional HOI categories rather than a single object identity. Its main objective is to adapt the model’s output to align more closely with the given subject, while maintaining the ability to generate diverse results. The loss function of DreamBooth consists of two components: predicted noise residual loss and class-specific prior preservation loss, defined as:
\begin{multline}
\mathbb{E}_{\mathbf{x}, \mathbf{c}, \boldsymbol{\epsilon}, \boldsymbol{\epsilon'}, t} \Big[ w_t \|\hat{x}_\theta(\alpha_t \mathbf{x} + \sigma_t \boldsymbol{\epsilon}, \mathbf{c}) - \mathbf{x}\|_2^2 + \\
\lambda w_{t'} \|\hat{x}_\theta(\alpha_{t'} \mathbf{x}_{\text{pr}} + \sigma_{t'} \boldsymbol{\epsilon'}, \mathbf{c}) - \mathbf{x}_{\text{pr}}\|_2^2 \Big]
\end{multline}

where, \( \mathbf{x} \) represents the HOI image, \( \mathbf{c} \) is the functional text condition vector, \( \boldsymbol{\epsilon} \) and \( \boldsymbol{\epsilon'} \) are the noise terms, and \( t \) is the time step in the diffusion process. The parameters \( \alpha_t \) and \( \sigma_t \) regulate the noise levels, while \( \alpha_{t'} \) and \( \sigma_{t'} \) serve a similar purpose, and \( \mathbf{x}_{\text{pr}} \) represents a sample generated by the pre-trained diffusion model. The weight factors \( w_t \) and \( w_{t'} \) regulate the contributions of each loss term, and \( \hat{x}_\theta \) represents the diffusion model. The hyperparameter \( \lambda \) adjusts the relative importance of the two loss components. 
During fine-tuning, each functional-category-specific DreamBooth model is optimized for 2000 iterations using AdamW with a learning rate of $1\times10^{-5}$ and prior-preservation weight $\lambda=1$.
Through DreamBooth fine-tuning, we can obtain the HOI image guided by the functional text prompt.

\begin{figure}[htbp]
  \centering
  \includegraphics[width=0.45\textwidth]{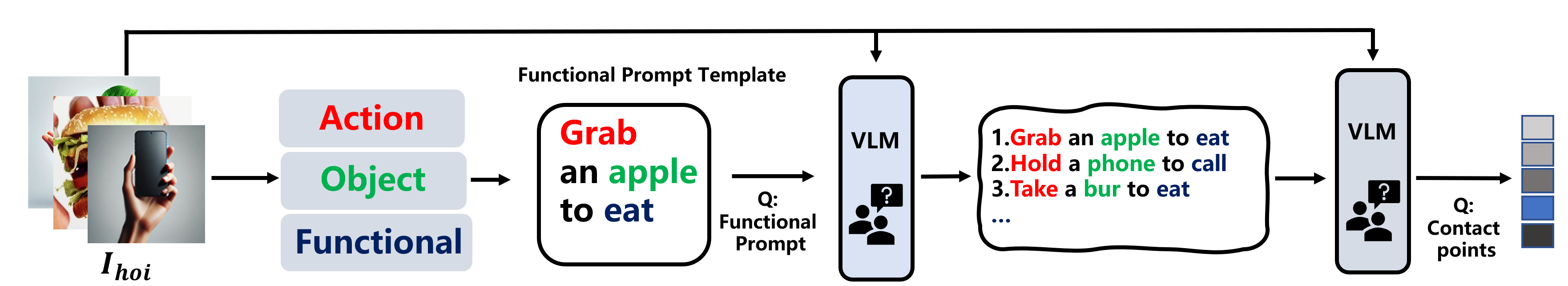} 
  \caption{Functional text and contact area generation process.}
  \label{fig:prompt}
\end{figure}

\textbf{Functional Prompt Generation}: We select valid contact frames from publicly available HOI datasets\cite{ho3D1,yang2022oakink} and drive LLaVA\cite{llava} to generate fine-grained functional descriptions by designing structured prompt templates: given the input HOI image $I_{hoi}$ and the question template "Describe the hand action, object function and usage purpose," the model outputs structured text $l$ such as 'Grab[action] an apple[object] to eat[task]'.  The detailed process is illustrated in Fig.~\ref{fig:prompt}.

\subsection{Geometric Reconstruction of hand and Object}\label{sec:3.1}

\textbf{Hand Representation and Reconstruction}: In order to reconstruct the hand mesh from the generated 2D HOI guidance image $I_{hoi}$ generated by the diffusion model. We use MANO~\cite{MANO:SIGGRAPHASIA:2017, hasson19_obman} as the hand representation for 3D HOI scenarios. MANO is a parametric model that provides a template mesh model of a standard hand \( M_{std}\), and models various appearances and poses of the hand using low-dimensional shape parameters \(\bm{\beta}\) and pose parameters \(\bm{\theta}\), allowing for the generation of corresponding hand mesh models $M_h = \text{MANO}(\bm{\theta}, \bm{\beta};M_{std})$. 

Optimizing HOI in 2D is challenging while validating its effectiveness is difficult. To obtain more refined and practical guidance HOI image \( I_{hoi} \in \mathbb{R}^{H \times W \times 3} \), we utilized HaMeR~\cite{pavlakos2024reconstructing} to elevate the 2D output of the diffusion model to 3D space. Adopting the denoised RGB image as input and given camera intrinsic parameters, HaMeR can predict the corresponding MANO parameters \(\bm{\theta}\) and \(\bm{\beta}\) for the hand, as well as the translation(denoted as \(\bm{t}\)) between the camera and hand, thereby the hand components of the 2D HOI can be elevated to 3D.

\textbf{Object Reconstruction from Single Image}: To recover the complete object image $I_{o}$, inspired by Affordance Diffusion\cite{ye2023affordance}, we adapt a diffusion-based approach to remove the hand information from \( I_{\text{hoi}} \). 

Specifically, we first extract the partially occluded object image from \( I_{\text{hoi}} \) and then input this image into the GLIDE\cite{glide} to inpaint the un-occluded object image. Finally, we utilize SAM\cite{sam} for segmentation to obtain the object image without the background, that used for the input of single image 3D reconstruction.

Single-image 3D reconstruction~\cite{clay,3Dshape,michelangelo,crastman} is a challenging task that involves generating a complete 3D model from a single image. To achieve this, we employ CLAY~\cite{clay}, an advanced generative 3D modeling system capable of producing high-fidelity meshes. 

CLAY leverages aligned shape-image latent representations as conditional inputs for a 3D generative model, enabling the efficient generation of detailed 3D shapes that are consistent with the reference image. Utilizing the object image as a conditional input, CLAY efficiently generates a congruent 3D mesh $M_{\text{o}}$:
\begin{equation}
M_{\text{o}} = D(E(I)+\epsilon)
\end{equation}
where, \( D \) represents the decoder, which reconstructs the 3D shape from the encoded noisy representation. \( E \) represents the encoder, which encodes the input object image \( I \)  into a latent noise space, and \( \epsilon \) is the noise. In this paper, the CLAY is responsible for reconstructing high-quality 3D mesh $M_{\text{o}}$ from the object images generated by diffusion.

\begin{figure}[htbp]
  \centering
  \includegraphics[width=0.5\textwidth]{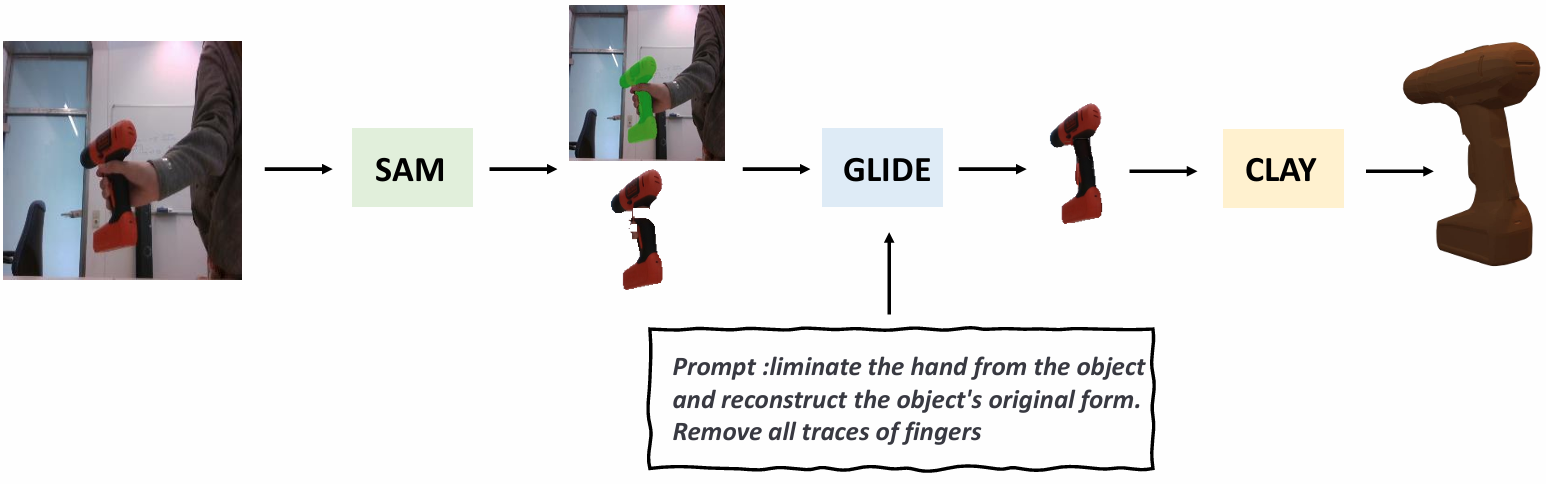} 
  \caption{Process for reconstructing 3D object meshes from HOI images.}
  \label{fig:objre}
\end{figure}

Fig.~\ref{fig:objre} illustrates the detailed process of object reconstruction. Initially, the HOI image is fed into the SAM model for object region segmentation, and the resulting segmentation is subsequently transferred to the GLIDE model. By leveraging meticulously designed prompts, the GLIDE model effectively supplements the object with detailed geometric information. Finally, the inpainted object image is input into the CLAY model to complete the 3D reconstruction, thereby yielding the final object mesh.

\textbf{Initial Scale Alignment of Hand-Object}.
We independently obtain the 3D hand $M_h$ and the object model $M_{\text{o}}$ from HaMeR and CLAY. However, a significant discrepancy exists in the scale ratio between the hand and object when compared to the 2D representations. To address this issue, we perform hand-object detection on the 2D HOI $I_{hoi}$ output from the diffusion model, yielding hand detection boxes \( \mathbf{B}_h \) and object detection boxes \( \mathbf{B}_o \). Based on this, we further calculate the hand-object scale ratio  \( k_{2D} \) in the $I_{hoi}$  by computing areas of boxes \( S(\mathbf{B_h}) \)  and \(S(\mathbf{B_o})\) :
\begin{equation}
k_{2D} = \sqrt{\frac{S(\textbf{B}_h)}{S(\textbf{B}_o)}}
\end{equation}
In the 3D space, we calculate the convex hulls \( \mathbf{C}_h \) and \( \mathbf{C}_o \) of the hand and object meshes respectively, and determined the volumes of these convex hulls $V(\mathbf{C}_h)$ and $V(\mathbf{C}_o)$ to derive the actual scale ratios \( k_{3D} \)  of the reconstructed 3D models:
\begin{equation}
k_{\text{3D}} = \sqrt[3]{\frac{V(\mathbf{C}_h)}{V(\mathbf{C}_o)}}
\end{equation}
By leveraging the scale of hand as a reference, we align the scale ratios to derive the scaling factor \( k_{scale} \) for $M_{\text{o}}$:
\begin{equation}
k_{\text{scale}} = \frac{k_{\text{3D}}}{k_{\text{2D}}}
\end{equation}
Finally, by scaling the $M_{\text{o}}$ according to this scaling factor, we obtain the object \(M_o = k_{\text{scale}} \cdot M_{\text{o}}\) of 3D HOI $M_{hoi}$ participants, which is consistent with the hand-object size ratio in the $I_{hoi}$.

\subsection{Pose Optimization Strategy}\label{sec:3.2}
In the Pose Optimization stage, our goal is to align the generated 3D object $M_o$ with the estimated hand $M_h$ and reconstruct the 3D HOI scene that is consistent with the generated functional guidance HOI image $I_{hoi}$.


\textbf{Object Pose Approximator}: Given the hand parameters $\{\bm{\theta} ,\bm{\beta}\}$, the  transformation from hand to camera $[\bm{R}| \bm{t}]_{\text{h} \rightarrow \text{c}}$, segmentation of the object $I_{mo}$, generated guidance HOI image $I_{hoi}$ and the reconstructed 3D object $M_o$. We first set the scale and translation of the object for initialization. We resize the object according to the 2d areas of hand and object detection boxes in $I_{hoi}$ and the volumes of their 3D models' convex hulls. We also set object's location to the center of the estimated hand to prevent it from being too far from the camera frustum.  After initialization, we optimize the transformation of the object $[\bm{R} | \bm{t}]_{\text{o} \rightarrow \text{c}}$ by ensuring the coverage between the rendered object and its corresponding segmentation. First, we sample $N$ points $P_o$ on the object surface using farthest-point sampling. Then we project the sampled surface points of the object onto the camera plane:
\begin{equation}
P_{surface}^{2d} = \left\{ \Pi(\bm{K})[\bm{R} | \bm{t}]_{\text{o} \rightarrow \text{c}}| \bm{x} \in P_{surface} \right\}
\end{equation}
where $\Pi(K)$ is the projection operation according to the camera intrinsic $K$, $P_{surface}$ is the set of object's sampled surface points. The camera is fixed at $[0,0,0]$ with the view matrix set to the identity matrix $\bm{I}$. We then compute the Chamfer Distance~\cite{chamferdis} between projected surface points $P_{surface}^{2d}$ and the object segmentation $I_{mo}$:
\begin{equation}
\mathcal{L}_{cd} = \sum_{\bm{x} \in P_{surface}^{2d}} \min_{\bm{y}} \| \bm{x} - \bm{y} \|^2 + \sum_{\bm{y} \in I_{mo}} \min_{\bm{x}} \| \bm{x} - \bm{y} \|^2
\end{equation}

By decreasing the $\mathcal{L}_{cd}$, the contour of the object will be aligned as closely as possible to the segmentation of the object $I_{mo}$. 

\textbf{Distance and Scale Selection}: Although the aforementioned alignment ensures that the contour of the object in camera plane is aligned with the segmented mask, the inaccurate object scale could lead to incorrect distance between the object and the camera. For example, when the object appears larger than its actual size, the method will shift the object farther from camera to ensure the mask aligns properly. Thereby, we apply a selection for the object's distance and scale. Given the object $M_o'$ which is aligned with the object segmentation, we first generate $N$ candidate objects, whose positions are distributed along center of object $M_o'$. The generated object sizes are adjusted according to the distance between the object and camera:
\begin{equation}
    \hat{M_o^i}=\{\frac{d_i}{||\bm{x}_c||}(\bm{x}-\bm{x}_c)+\frac{d_i}{||\bm{x}_c||}\bm{x}_c |\bm{x}\in M_o'\}
\end{equation}
where $\hat{M_o^i}$ is the $i$-th generated object, $d_i$ is the sampled distance and $\bm{x}_c$ is the object center of $M_o'$. Then we select the object with the smallest distance to the hand contact points from the $N$ generated candidates, denoted as $\hat{M_o}$, which will be used in the subsequent optimization with fixed transformation.

\textbf{Hand Contact Optimization}: 
To analyze the contact regions, we design a visual question answering (VQA) task with the prompt ``Identify the contact areas between the hand and the object.'' We feed the HOI image into a Vision-Language Model (VLM) and query: ``Which fingers are in contact with the object in this HOI image?'' Since the hand keypoints in the MANO hand model are predefined, once the finger contact information is obtained, we use this information along with the corresponding MANO hand keypoints as the basis for the hand-object contact optimization. The detailed process is illustrated in Fig.~\ref{fig:prompt}.

\begin{figure}[htbp]
  \centering
  \includegraphics[width=0.5\textwidth]{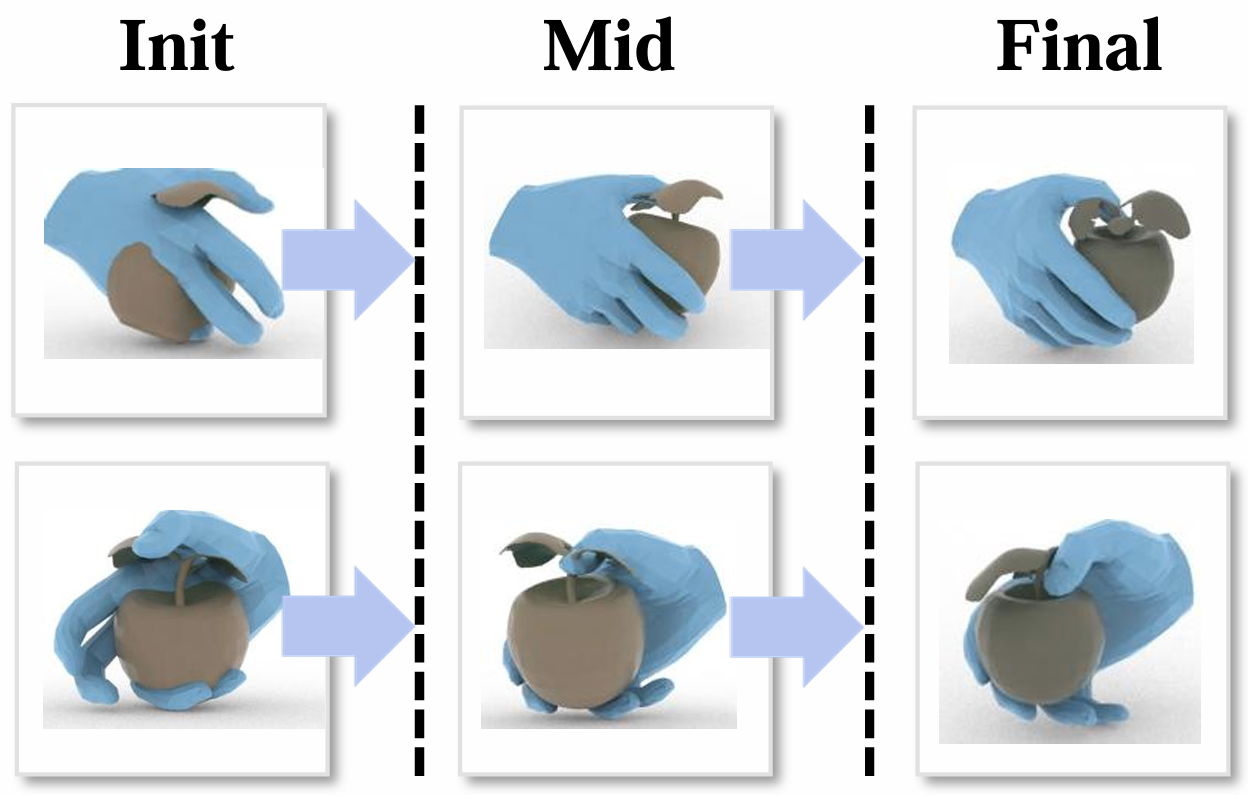} 
  \caption{The hand-object contact optimization refines originally interpenetrating hands and object into a physically plausible HOI 3D scene.}

  \label{fig:hco}
\end{figure}

To further address the imperfections in the generated HOI scenes—such as floating and interpenetration during hand-object contact—we draw inspiration from several energy terms proposed in DexGraspNet\cite{dexgraspnet} and design a hand contact optimization module to enhance the physical plausibility and quality of the interaction. We first define the following distance loss $\mathcal{L}_{dis}$ to ensure the hand contact points are as close as possible to the object surface:
\begin{equation}
\mathcal{L}_{dis} = \sum_{\bm{x}\in P_{contact}}|SDF(x,M_o)|
\end{equation}
where $P_{contact}$ are the contact points on the hand, which are determined by the VLM.
$SDF(\cdot,M_o)$ is the signed distance function of object $M_o$.
We define the interpenetration loss to avoid interpenetration between the hand and the object:
\begin{equation}
\mathcal{L}_{pen} = \sum_{\bm{x}\in P_{contact}} -min(SDF(x,M_o),0)
\end{equation}
where the value of $SDF(x,M_o)$ is negative when $x$ is inside the object $M_o$.
We further introduce the self-penetration loss from DexGraspNet \cite{dexgraspnet} to ensure the plausibility of optimized hand pose:
\begin{equation}
\mathcal{L}_{spen} = \sum_{\bm{x}\in P_{hand}}\sum_{\bm{y}\in P_{hand}}[x\neq y] max(\delta-||\bm{x}-\bm{y}||, 0)
\end{equation}
where $P_{hand}$ are the surface points of hand model.
To prevent the optimized hand pose from deviating too much from the estimated HaMeR parameters, we propose a supervision loss:
\begin{equation}
\mathcal{L}_{sup} = ||\hat{\theta}-\theta||
\end{equation}
where the $\hat{\theta}$ is the optimized hand parameters and the $\theta$ is the estimated hand parameter from HaMeR.
Overall, our optimization objective is to minimize the following loss function:
\begin{equation}
\mathcal{L} = \mathcal{L}_{dis} + \lambda_{pen} \mathcal{L}_{pen} + \lambda_{spen} \mathcal{L}_{spen} + \lambda_{sup} \mathcal{L}_{sup}
\end{equation}
where $\lambda_{pen}$, $\lambda_{spen}$, $\lambda_{sup}$ are hyperparameters. The hand contact optimization process can be referred to in Fig.~\ref{fig:hco}.
%

\section{Experiments}
In this section, we evaluate FunHOI from multiple perspectives. 
Sec.~\ref{sec:4.1} describes the implementation details and experimental setup. 
Sec.~\ref{sec:4.2} introduces the evaluation metrics and datasets. 
Sec.~\ref{sec:4.3} compares FunHOI with representative 3D HOI reconstruction and text-guided interaction generation methods. 
Sec.~\ref{sec:4.4} analyzes the contribution of individual components through ablation studies.

Since FunHOI targets functional-intent-conditioned 3D HOI synthesis without requiring 3D HOI annotations, there is no directly matched benchmark with the same input, supervision, and output setting.
We therefore evaluate the framework under two complementary protocols.
First, to quantify reconstruction quality with available 3D ground truth, we replace the generated 2D HOI guidance image $I_{hoi}$ with images from public HOI datasets and compare the reconstructed 3D hand--object interactions with existing image-based reconstruction methods.
Second, to assess functional text guidance and physical plausibility, we compare FunHOI with recent text-guided and video-based HOI baselines under their corresponding input assumptions, as detailed in Sec.~\ref{sec:4.3}.

\subsection{Implementation Details}\label{sec:4.1}

All experiments are performed on a system with a single NVIDIA GeForce RTX 4090 GPU. 

For Functional Grasp Generation, the diffusion model is fine-tuned in a functional-category-specific manner rather than jointly over all interaction categories.
We collect 1000 valid HOI images from HO3D and OakInk, covering 30 functional interaction categories, with approximately 30--50 images per category depending on the availability of valid contact frames.
Each image is paired with an action--object--purpose prompt generated following Sec.~\ref{sec:p}.
For each functional category, a separate DreamBooth model is fine-tuned for 2000 iterations using AdamW with learning rate $1 \times 10^{-5}$ and prior-preservation weight $\lambda=1$.
During inference, classifier-free guidance is applied with scale $s_{\mathrm{cfg}}=7.5$.

In the Functional Grasp Refiner, we use the Adam~\cite{adam} optimizer for all optimization tasks. 
The hyperparameter settings for FGR are summarized in Tab.~\ref{tab:hyperparameters}. 
The same FGR hyperparameters are used across datasets without target-dataset-specific tuning.
Generating a 2D HOI guidance image takes approximately 4~s and 2.7~GB GPU memory.
Hand and object mesh reconstruction require about 46~s/5.8~GB and 24~s/14.7~GB, respectively.
The hand--object composition and contact optimization stage takes approximately 3~min and 1.1~GB GPU memory.
DreamBooth fine-tuning for each functional interaction category takes around 10~min with 16--18~GB peak memory.
DreamBooth fine-tuning is performed once per functional interaction category and is excluded from the per-sample inference time.
Excluding fine-tuning, the overall inference time is approximately 4--5~min per sample.
The detailed DreamBooth fine-tuning and inference settings are summarized in Tab.~\ref{tab:DreamBooth_hyper}.

\begin{table}[t]
\centering
\caption{DreamBooth fine-tuning and inference settings.}
\label{tab:DreamBooth_hyper}
\scriptsize
\setlength{\tabcolsep}{2.6pt}
\renewcommand{\arraystretch}{1.05}
\resizebox{\columnwidth}{!}{
\begin{tabular}{lccccccccccc}
\toprule
\textbf{Strat.} &
\textbf{Img.} &
\textbf{Opt.} &
\textbf{LR} &
\textbf{Iter.} &
\textbf{Res.} &
\textbf{BS} &
\textbf{Prior} &
\textbf{Cls.} &
\textbf{Sch.} &
\textbf{Warm.} &
\textbf{CFG} \\
\midrule
Per-cat. &
30--50 &
AdamW &
$1{\times}10^{-5}$ &
2000 &
$512^2$ &
4 &
$\lambda{=}1$ &
5 &
Const. &
0 &
7.5 \\
\bottomrule
\end{tabular}
}
\end{table}

\begin{table}[htbp]
  \centering
  \caption{Hyperparameters for Object Pose Approximator (OPA) and Contact Optimization.}
  \label{tab:hyperparameters}
  \begin{tabular}{@{}lcc@{}}
    \toprule
    Stage & Parameter & Value \\
    \midrule
    \multirow{5}{*}{\centering \textbf{Object Pose Approximator}} 
    & \( \lambda_{\text{cam}} \) & 1000 \\
    & \( \lambda_{\text{dep}} \) & 0.1 \\
    & iterations & 200 \\
    & \( \text{learning rate} \)  & \( 1 \times 10^{-3} \) \\
    & \( N \)  & \( 1000 \) \\
    \midrule
    \multirow{7}{*}{\centering \textbf{Hand Contact Optimization}} 
    & \( \lambda_{pen} \)& 1 \\
    & \( \lambda_{spen} \) & 0.75 \\
    & \( \lambda_{sup} \) & 0.3 \\
    & iterations & 2000 \\
    & \( lr_{translate} \) & \(1 \times 10^{-4}\) \\
    & \( lr_{rotation} \) & \(1 \times 10^{-3}\) \\
    & \( lr_{axis} \) & \(8 \times 10^{-4}\) \\
    \bottomrule
  \end{tabular}
\end{table}
\begin{figure*}[htbp]
  \centering
  \includegraphics[width=1.0\textwidth]{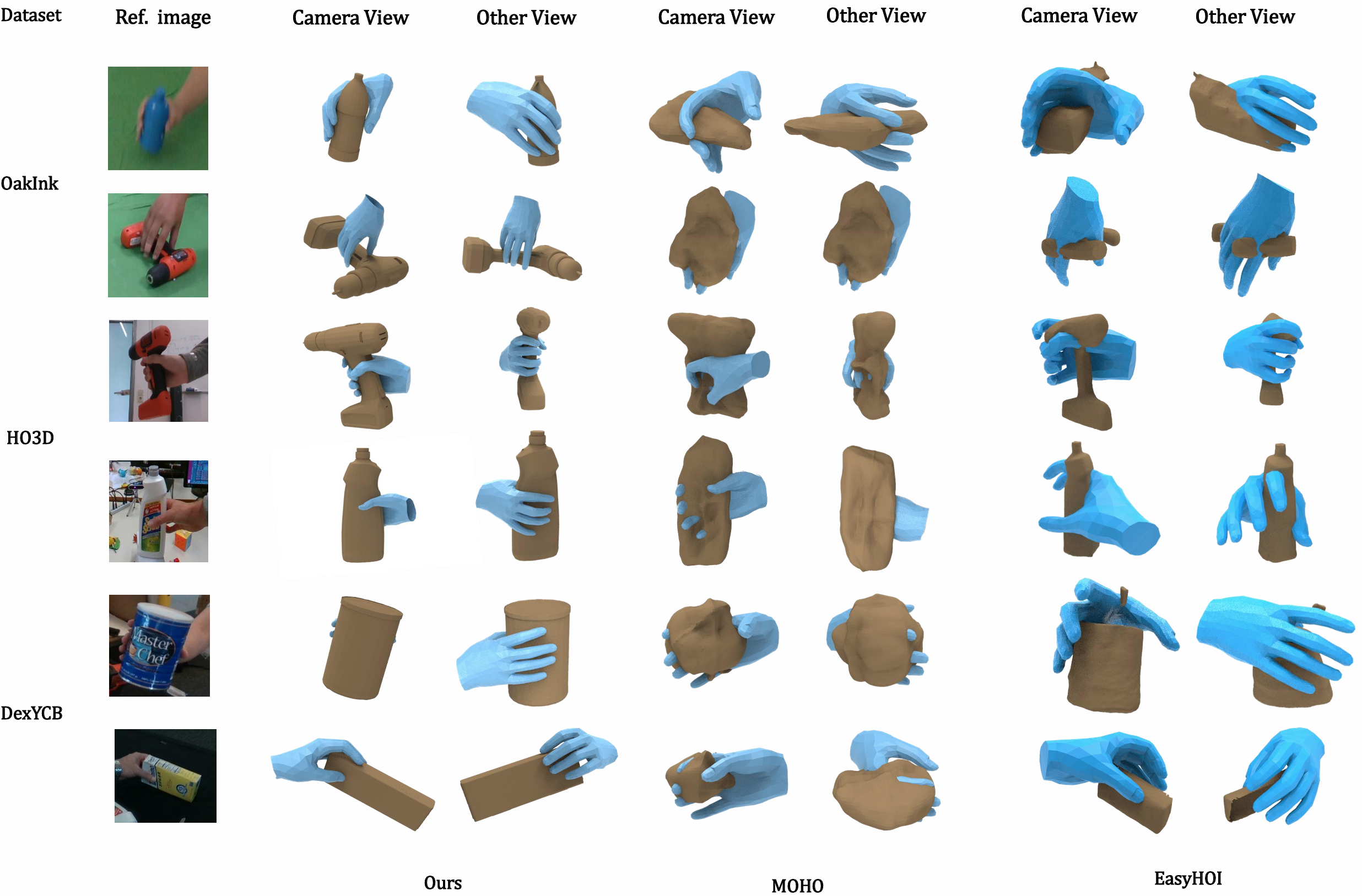}  
  \caption{3D Reconstruction Visualization of FunHOI on the OakInk, HO3D, and DexYCB Datasets. The first column shows the input images. Compared with other methods, FunHOI produces visually coherent hand--object configurations with improved contact plausibility. Additional visualizations are provided in the appendix.}
  \label{fig:res}
\end{figure*}
\subsection{Metrics and Dataset}\label{sec:4.2}

\textbf{Metrics.} 
To assess reconstruction quality, we compute the F-score at 5~mm and 10~mm thresholds, which balances precision and recall and provides a robust measure of surface alignment. 
We also employ Chamfer Distance (CD) to quantify the average bidirectional discrepancy between reconstructed and ground-truth meshes. 
To evaluate the physical plausibility of hand--object interactions, we measure the Solid Intersection Volume (SIV) between the reconstructed hand and object meshes, which penalizes interpenetration artifacts.

For evaluating the contribution of different modules, we use Simulation Displacement (SD), Solid Intersection Volume (SIV), and Contact Ratio (CR). 
Because SIV and CR alone may not fully characterize interaction quality, we additionally report the Interaction Plausibility Index (IPI), defined as $IPI = CR / \exp(SIV)$. 
IPI jointly reflects the degree of valid contact and the amount of interpenetration, providing a complementary measure of hand--object interaction plausibility.

\textbf{Dataset.}
The HO-3D dataset~\cite{ho3D1} provides 3D pose annotations for hand--object interaction scenarios with severe occlusion. 
It consists of sequences in which different subjects interact with objects selected from the YCB dataset~\cite{ycb}. 
OakInk~\cite{yang2022oakink} is a large-scale multimodal dataset for visual and cognitive understanding of HOI. 
It includes Oak, a functional knowledge graph based on 1800 common household objects, and Ink, which contains human interaction records for 100 objects. 
DexYCB~\cite{chao2021dexycb} is a high-quality RGB-D dataset for hand--object interaction, containing 582K frames from 1000 sequences of 20 objects grasped by 10 subjects from 8 viewpoints.

We randomly select 500 images from each dataset for evaluation. 
The supplementary material provides additional HOI reconstruction results for representative categories. 
Since IHOI and gSDF were trained on HO3D and DexYCB, respectively, we exclude their results on the corresponding training datasets to avoid train-test overlap.

\begin{table*}[t]
\centering
\caption{
Evaluation protocol and key differences from representative baselines.
We summarize each method in terms of input modality, object source, hand source, 3D HOI annotation requirement, 3D HOI training dependency, video dependency, and evaluation setting.
}
\label{tab:baseline_protocol}
\vspace{0.3em}
\scriptsize
\setlength{\tabcolsep}{4.0pt}
\renewcommand{\arraystretch}{1.08}
\resizebox{\textwidth}{!}{
\begin{tabular}{@{}llllcccl@{}}
\toprule
\textbf{Method} &
\textbf{Input} &
\textbf{Object Source} &
\textbf{Hand Source} &
\textbf{3D HOI Anno.} &
\textbf{3D HOI Train.} &
\textbf{Video} &
\textbf{Eval. Setting} \\
\midrule
IHOI~\cite{ihoi} &
RGB image &
Reconstructed &
Estimated &
\cmark &
\cmark &
\xmark &
Same images \\
gSDF~\cite{gsdf} &
RGB image &
Implicit SDF &
Estimated &
\cmark &
\cmark &
\xmark &
Train set excluded \\
HOLD~\cite{hold} &
Video/keyframe &
Optimized &
Optimized &
\cmark &
\cmark &
\cmark &
Contact frames \\
Text2HOI~\cite{t2h} &
Text + template &
Object template &
Generated &
\cmark &
\cmark &
\xmark &
Same text/template \\
DiffH2O~\cite{christen2024diffh2o} &
Text + obj. geometry &
Object template &
Generated &
\cmark &
\cmark &
\xmark &
Same prompts/geometry \\
Text2Grasp~\cite{text2grasp} &
Text + obj. point cloud &
Object template &
Generated grasp &
\cmark &
\cmark &
\xmark &
Same prompts/geometry \\
\textbf{FunHOI (Ours)} &
\textbf{Text / RGB image} &
\textbf{Single-image recon.} &
\textbf{Estimated} &
\textbf{\xmark} &
\textbf{\xmark} &
\textbf{\xmark} &
\textbf{Same split; test-time opt.} \\
\bottomrule
\end{tabular}
}
\end{table*}

\subsection{Comparison with State-of-the-Art Methods}\label{sec:4.3}
\textbf{Comparison Protocol.}
The compared methods operate under different input and supervision assumptions. To make the evaluation setting explicit, Tab.~\ref{tab:baseline_protocol} summarizes the input modality, object source, hand source, 3D HOI annotation requirement, 3D HOI training dependency, video dependency, and evaluation protocol for each method. For text-guided baselines such as Text2HOI, DiffH2O, and Text2Grasp, we use the same prompts and object geometry or templates whenever applicable. For image- and video-based reconstruction methods, we follow their required input settings and exclude training-set overlaps where necessary. This protocol separates differences in task formulation from differences in empirical performance.

\begin{table*}[!t]
\centering
\small
\setlength{\tabcolsep}{3pt} 
\newcolumntype{C}{>{\centering\arraybackslash}X} 
\begin{tabularx}{\textwidth}{lCCCCCCCCCCCC} 
\toprule
                      & \multicolumn{4}{c}{\textbf{OakInk\cite{yang2022oakink}}}    & \multicolumn{4}{c}{\textbf{HO3D\cite{ho3D1}}}  & \multicolumn{4}{c}{\textbf{DexYCB\cite{chao2021dexycb}}}  \\ 
                      \cmidrule(lr){2-5} \cmidrule(lr){6-9} \cmidrule(lr){10-13}
                      & {F$_{5}\uparrow$} & {F$_{10}\uparrow$} & {CD$\downarrow$} & {SIV$\downarrow$} & {F$_{5}\uparrow$} & {F$_{10}\uparrow$} & {CD$\downarrow$} & {SIV$\downarrow$} &   {F$_{5}\uparrow$} & {F$_{10}\uparrow$} & {CD$\downarrow$} & {SIV$\downarrow$} \\
\midrule
IHOI\cite{ihoi} & 0.097  & 0.152  & 1.742 & 0.0466 & $-$ & $-$ & $-$ & $-$ & 0.084 & 0.143 & 1.897 & 0.0493 \\
gSDF\cite{gsdf} & 0.106	&0.173  &  1.992 	& 0.0416  & 0.231  & 0.414  &  1.284 &	0.0432 & $-$	& $-$&   $-$ &  $-$  \\
MOHO\cite{zhang2024moho} & 0.175 	&  0.323 	& 3.883 	& 0.0486	&0.210  	&0.378 	&1.393	&0.0476 & 0.119 &0.249	& 1.695 & 0.0461	\\
EasyHOI\cite{liu2024easyhoi} & 0.247 &0.445 &  1.035 	& 0.0411 &	0.243 &	0.432 &  1.204	& 0.0432  &	0.134  &  0.253  &1.628 & 0.0452\\
\midrule
\textbf{FunHOI (Ours)}  & \textbf{0.250}   & \textbf{0.446}  & \textbf{1.088} &\textbf{0.0093} & \textbf{0.317} & \textbf{0.537} &\textbf{1.048} & \textbf{0.0064} & \textbf{ 0.218} & \textbf{0.392} & \textbf{1.330} & \textbf{0.0092} \\
\bottomrule
\end{tabularx}
\caption{The evaluation results for 3D HOI reconstruction are presented, using $F_{5}$ ($mm$), $F_{10}$ ($mm$), Chamfer Distance ($mm$), and Solid Intersection Volume ($\times 100cm^3$) as metrics. Overall, FunHOI significantly outperforms the other baselines.
}
\label{tab:result}
\end{table*}

\textbf{Visualization Results.}
Fig.~\ref{fig:res} visualizes the comparison between FunHOI and image-based 3D HOI reconstruction methods. 

Compared with existing approaches, FunHOI produces hand-object configurations that are more consistent with the reference image in terms of object scale, contact region, and relative hand-object placement. These qualitative results support the effectiveness of the proposed scale-aware alignment and contact refinement strategy.

\begin{figure}[htbp]
  \centering
  \includegraphics[width=0.5\textwidth]{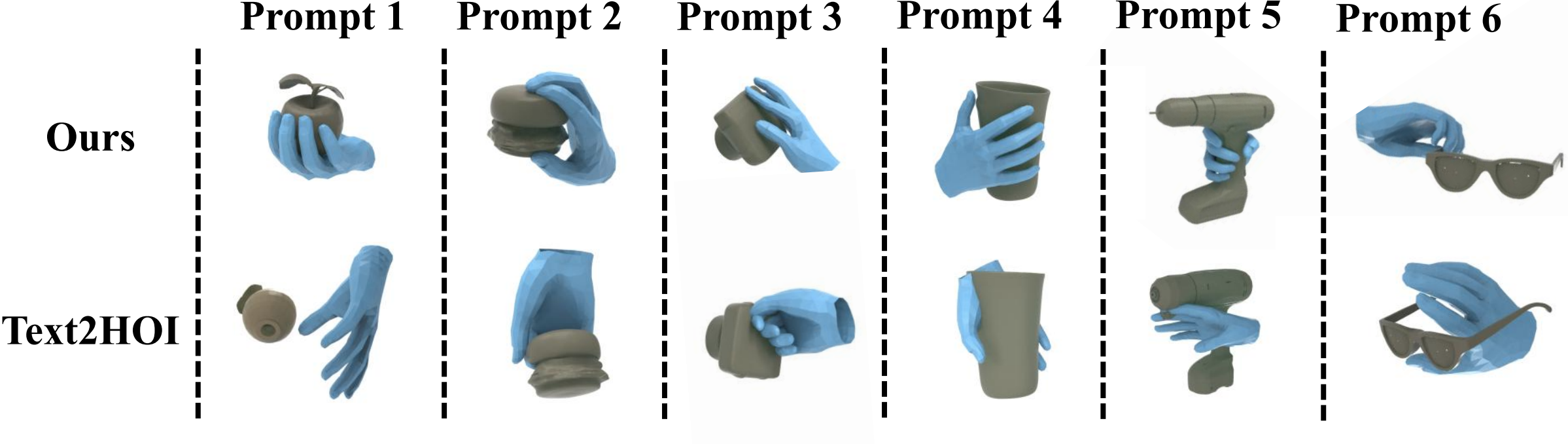} 
  \caption{Compared with Text2HOI\cite{t2h}, under the same textual conditions and using identical object templates, our method achieves more faithful reconstruction of 3D HOI scenes that align with human semantic understanding.}
  \label{fig:t2h}
\end{figure}

To further validate the advantages of our framework, we provide a visual comparison with Text2HOI\cite{t2h}. The object templates required by Text2HOI are reconstructed using CLAY. Fig.~\ref{fig:t2h} demonstrates that our method generates more plausible HOI scenes compared to Text2HOI.

\begin{figure}[htbp]
  \centering
  \includegraphics[width=0.45\textwidth]{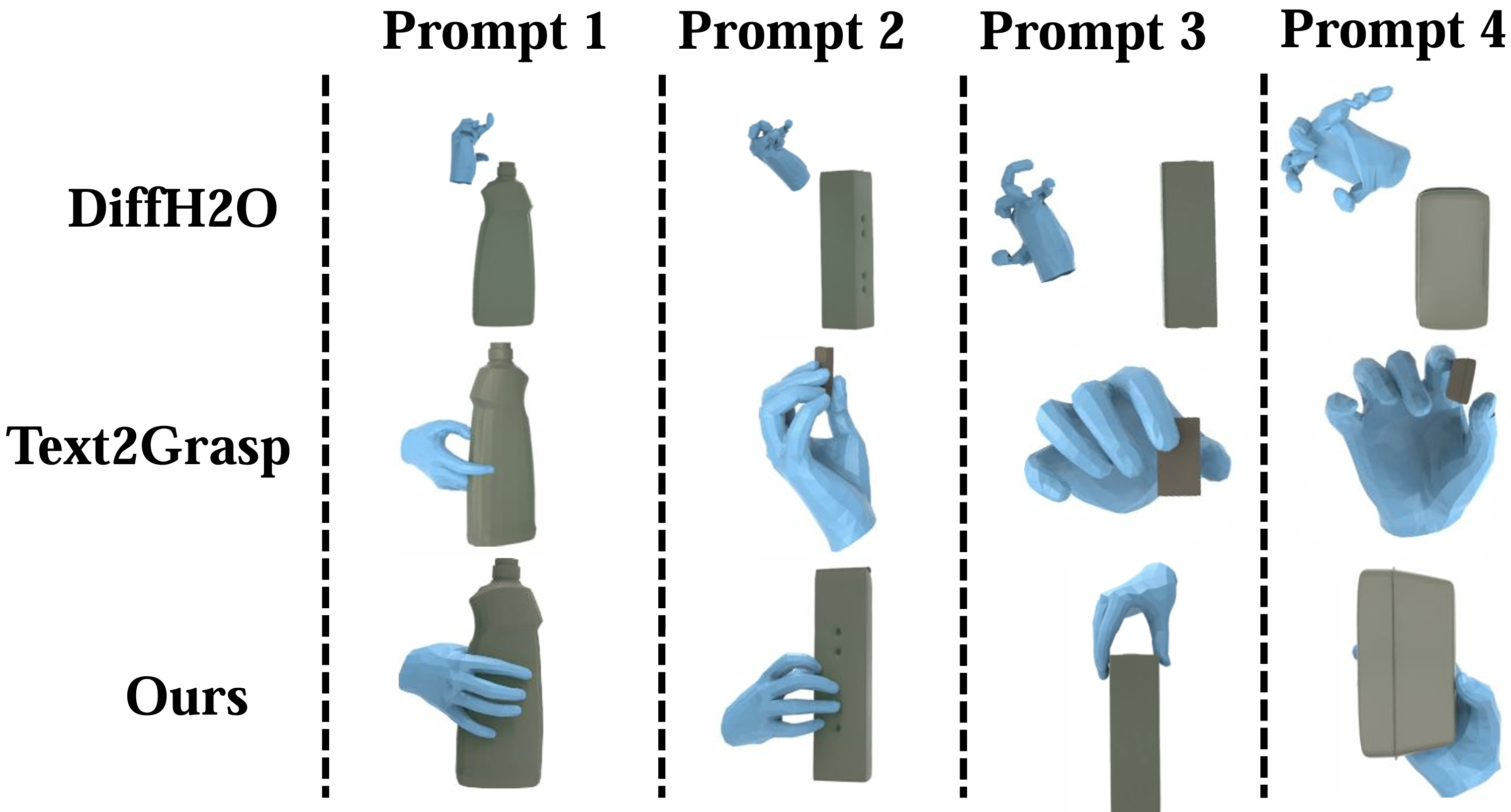}  
  \caption{
Comparison with text-guided HOI generation baselines (DiffH2O and Text2Grasp) under multiple functional prompts. Our method produces more semantically aligned and geometrically consistent 3D hand-object interactions, while avoiding unstable hand poses and unrealistic object placements observed in prior approaches.
}

  \label{fig:text}
\end{figure}

Moreover, we evaluate our method in the text-driven generation setting against DiffH2O\cite{christen2024diffh2o} and Text2Grasp\cite{text2grasp}. As illustrated in Fig.~\ref{fig:text}, DiffH2O tends to generate unstable hand poses and insufficient hand-object contact when encountering objects outside its training distribution, resulting in distorted or implausible interaction configurations. Text2Grasp, while producing reasonable grasp poses, often introduces noticeable object-scale distortions and lacks geometric consistency in the reconstructed 3D scene. Furthermore, both methods rely on large-scale supervised training to learn text-conditioned grasp priors, which limits their scalability and generalization to unseen domains. In contrast, FunHOI does not require supervised 3D HOI training data or 3D HOI annotations, and operates directly from a single HOI image, enabling robust and semantically coherent 3D HOI reconstruction even for out-of-distribution objects.

\begin{figure}[htbp]
  \centering
  \includegraphics[width=0.5\textwidth]{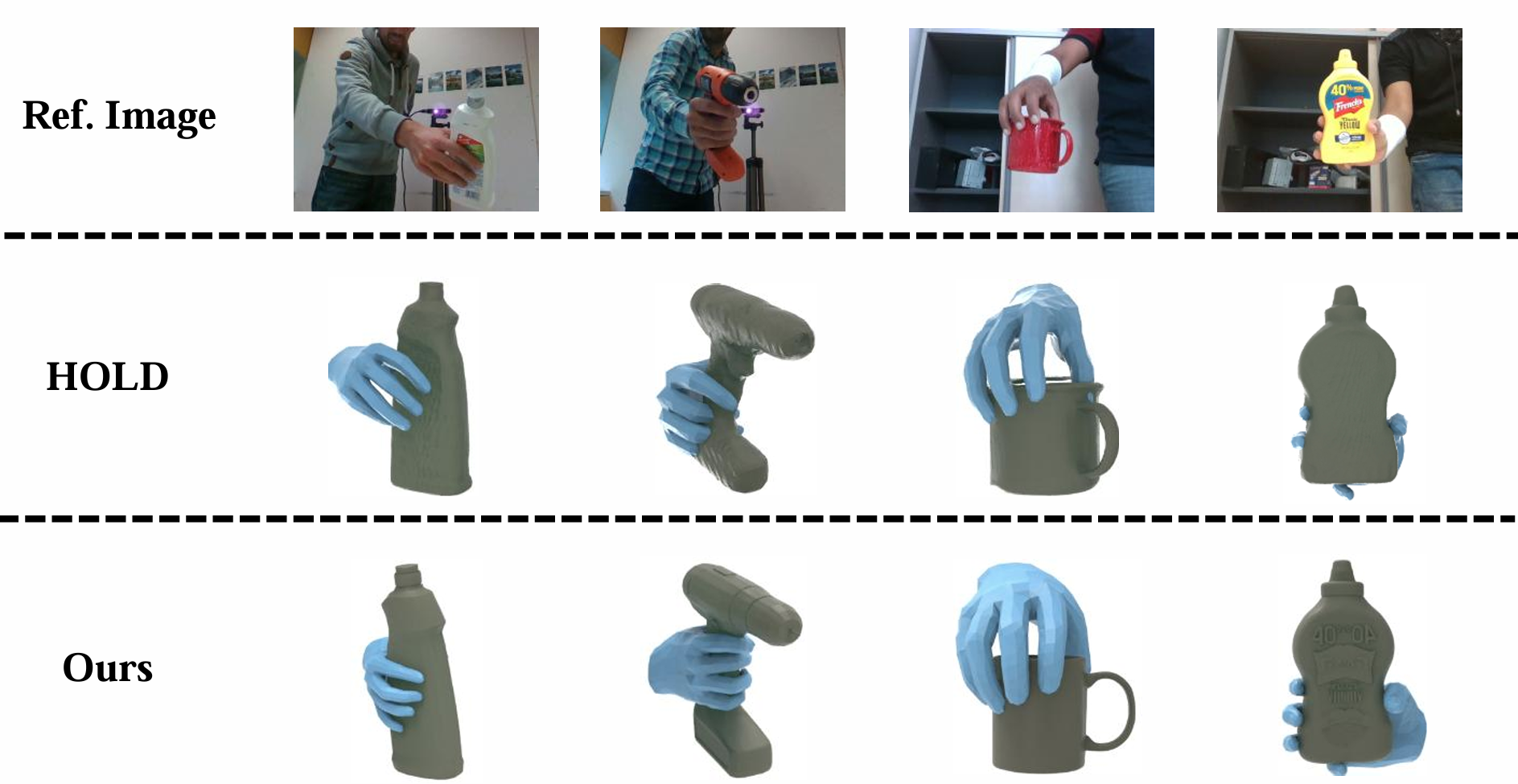}  
  \caption{
 Comparison with HOLD. Under identical reference images, our method produces significantly improved physical plausibility, reducing hand--object interpenetration and achieving more stable and realistic contact configurations. These qualitative observations are consistent with the quantitative improvements in SIV, Mean Penetration Depth (MPD), Max Penetration Depth (MaxPD), and Contact Ratio (CR) reported in Table~\ref{tab:physical_eval}.
  }
  \label{fig:hold}
\end{figure}

Moreover, we compare our method with HOLD\cite{hold} on HO3D, using the pre-trained HOLD model and selecting keyframes with clear hand-object contact for fair evaluation. Under identical reference images, our approach achieves comparable reconstruction fidelity while substantially improving physical interaction quality. As illustrated in Fig.~\ref{fig:hold}, our method produces more stable hand-object configurations with visibly reduced interpenetration and more coherent contact regions. Quantitatively,Table~\ref{tab:physical_eval} shows that our method reduces interpenetration volume by nearly 9$\times$, significantly lowers both the mean and maximum penetration depths, and doubles the contact ratio compared to HOLD, indicating substantially improved physical plausibility. Notably, HOLD relies on dataset-specific training and supervised video data, limiting its generalization to out-of-domain scenarios. In contrast, FunHOI does not require supervised 3D HOI training data or 3D HOI annotations, and operates directly on a single HOI image, enabling scalable and flexible application to unseen categories and open-domain scenarios.

\begin{table}[t]
\centering
\caption{
Physical Interaction Evaluation.
Lower SIV, Mean Penetration Depth (MPD), and Max Penetration Depth (MaxPD) indicate less interpenetration,
while higher Contact Ratio (CR) indicates better hand--object contact.
All depth-based metrics are averaged over two sequences.
}
\begin{tabular}{lcccc}
\hline
Method & SIV $\downarrow$ (m$^3$) & MPD $\downarrow$ (m) & MaxPD $\downarrow$ (m) & CR(\%) $\uparrow$ \\
\hline
HOLD & 3.24e-4 & 0.032 & 0.084 & 0.036 \\
\textbf{Ours} & \textbf{3.63e-5} & \textbf{0.016} & \textbf{0.078} & \textbf{0.072} \\
\hline
\end{tabular}
\label{tab:physical_eval}
\end{table}

\begin{figure}[htbp]
  \centering
  \includegraphics[width=0.5\textwidth]{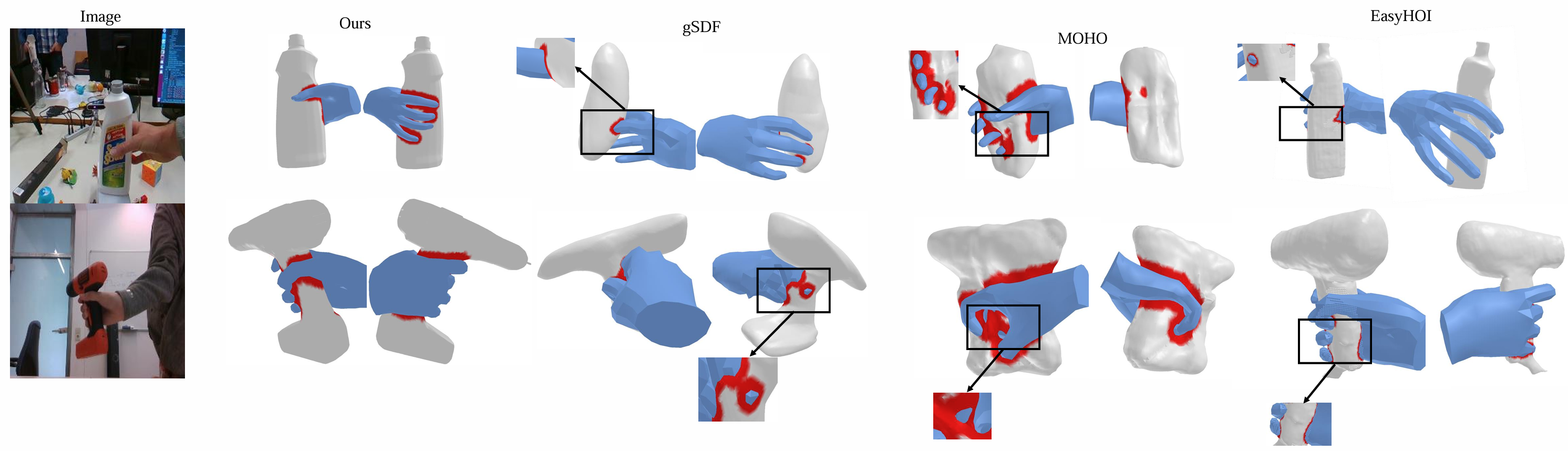} 
  \caption{Contact area visualization comparing our method with other 3D HOI reconstruction approaches. Our method improves upon existing baselines in both reconstruction quality and the reduction of interpenetration issues.}
  \label{fig:chuanmo}
\end{figure}

Meanwhile, Fig.~\ref{fig:chuanmo} further illustrates the superiority of our method in addressing hand-object penetration issues, highlighting its effectiveness in ensuring accurate and realistic HOI. 
In addition, we fine-tune the diffusion model on various web images. The results clearly demonstrate that our method can effectively generate diverse and high-quality 3D HOI scenes. The visualized results are presented in the supplementary material.

\textbf{Quantitative Evaluation.} Tab.~\ref{tab:result} presents the quantitative results of our method in comparison with other approaches. The results indicate a significant improvement in reconstruction quality over current state-of-the-art models. Additionally, due to the application of an energy function that guides the optimization of hand-object contact, our method outperforms other algorithms by a substantial margin in terms of the Solid Intersection Volume (SIV) metric.

\textbf{Cross-Dataset Transfer.}
To evaluate cross-dataset generalization, we fine-tune FGG on one dataset and evaluate the generated 3D HOI results on another. FGR is kept fixed, and its hyperparameters are not tuned on the target dataset. Since single-view HOI reconstruction has inherent global scale and translation ambiguity, we align each prediction to the ground-truth bounding-box center and diagonal before computing F5, F10, and CD. This protocol focuses the evaluation on relative hand-object geometry and interaction quality rather than global coordinate offsets.
As reported in Table~\ref{tab:cross_dataset}, FunHOI maintains reasonable transfer performance in both directions. The HO3D target setting is more challenging, likely due to its smaller object diversity and more constrained interaction distribution. These results indicate that the proposed FGG--FGR pipeline is not restricted to a single dataset-specific distribution.

\begin{table}[t]
\centering
\caption{
Cross-dataset transfer evaluation. The FGG model is fine-tuned on the source dataset and evaluated on the target dataset. Predictions are aligned to the ground-truth bounding-box center and diagonal before computing F5/F10/CD to account for monocular global scale and translation ambiguity.
}
\label{tab:cross_dataset}
\begin{tabular}{l l c c c}
\hline
Source & Target GT & F5 $\uparrow$ & F10 $\uparrow$ & CD $\downarrow$ \\
\hline
HO3D & OakInk & 0.242 & 0.427 & 1.359 \\
OakInk & HO3D & 0.144 & 0.273 & 3.054 \\
\hline
\end{tabular}
\end{table}

\begin{figure}[htbp]
  \centering
  \includegraphics[width=0.5\textwidth]{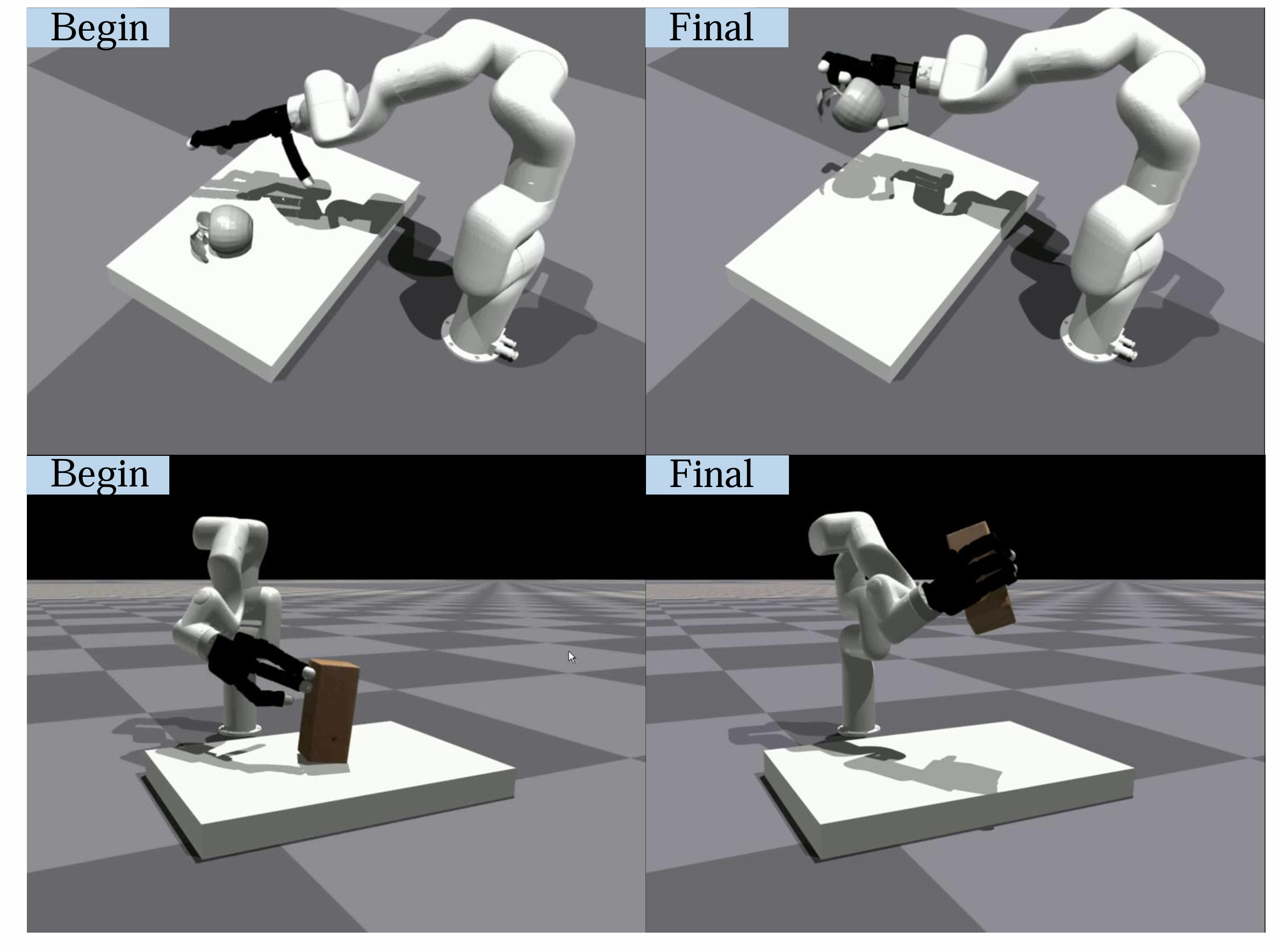} 
  \caption{Rendered view of the “grasp and lift” action executed by the trained policy using our generated 3D HOI.}
  \label{fig:dm}
\end{figure}

\textbf{Additional Experiments.} To further validate the effectiveness of our method, we integrate the generated 3D HOI into the dexterous manipulation simulation using the HOP\cite{hop}. As shown in Fig.~\ref{fig:dm}, the 3D models produced by our approach successfully perform dexterous manipulations. The corresponding demonstration videos are included in the supplementary material.


\textbf{User Study.} We conducted a user study with 30 participants to evaluate whether the generated hand--object interactions align with human interpretations of functional text descriptions. The study evaluated 50 generated 3D hand--object interaction results, covering a variety of functional grasping scenarios. For each functional prompt (e.g., ``grasp the mug to drink''), participants were presented with the 3D HOI results produced by different methods and were asked to select the single interaction that best satisfied the intended function described in the text, considering grasp purpose, hand--object relationship, and overall plausibility. For each sample, the method receiving the largest number of participant selections was counted as the preferred method for that sample. To avoid potential bias, the presentation order was randomized and participants were not informed of which method produced each result. We report a Top-1 Preference metric, defined as the percentage of samples for which a method is selected as the most preferred one across participants. As summarized in Table~\ref{tab:user_study}, FunHOI is selected as the most preferred method in a substantially larger portion of samples compared to baseline approaches, indicating better alignment with human functional understanding.

\begin{table}[t]
\centering
\caption{User study results on functional intent alignment.
We report the percentage of samples for which each method
receives the largest number of participant selections.}
\label{tab:user_study}
\begin{tabular}{l c}
\toprule
Method & Top-1 Preference (\%) $\uparrow$ \\
\midrule
IHOI     & 5.8 \\
gSDF     & 10.3 \\
MOHO    & 18.2 \\
EasyHOI & 20.5 \\
\textbf{FunHOI (Ours)} & \textbf{45.2} \\
\bottomrule
\end{tabular}
\end{table}


\subsection{Ablation Study}\label{sec:4.4}



\textbf{Controlled Integration Ablation.}
To isolate the contribution of the Functional Grasp Refiner, we conduct a controlled integration ablation where all variants use the same 2D guidance image, HaMeR hand reconstruction, and CLAY object reconstruction. Starting from a naive off-the-shelf integration baseline, we progressively add scale alignment, the Object Pose Approximator (OPA), and contact-aware refinement. As shown in Table~\ref{tab:controlled_integration_ablation}, the performance improves consistently as each component is introduced. Since the input hand and object reconstructions are fixed across all variants, the improvements directly reflect the effect of the proposed scale-aware and contact-aware composition strategy rather than the reconstruction backbones.

\begin{table*}[t]
\centering
\caption{
Controlled integration and FGR stage-wise ablation. All variants use the same 2D guidance image, HaMeR hand reconstruction, and CLAY object reconstruction. Starting from a naive off-the-shelf integration baseline, we progressively add scale alignment, OPA, and contact-aware refinement.
}
\label{tab:controlled_integration_ablation}
\resizebox{\textwidth}{!}{
\begin{tabular}{lcccccc}
\toprule
\textbf{Method} 
& \textbf{Scale} 
& \textbf{OPA} 
& \textbf{Contact Refine.} 
& \textbf{F@5} $\uparrow$ 
& \textbf{F@10} $\uparrow$ 
& \textbf{CD} $\downarrow$ \\
\midrule
Naive Integration 
& \xmark 
& \xmark 
& \xmark 
& 0.067
& 0.134
& 8.082 \\
+ Scale Alignment 
& \cmark 
& \xmark 
& \xmark 
& 0.082
& 0.159
& 7.657 \\
+ OPA 
& \cmark 
& \cmark 
& \xmark 
& 0.214 
& 0.317
& 1.566 \\
\textbf{Full FunHOI} 
& \cmark 
& \cmark 
& \cmark 
& \textbf{0.236}
& \textbf{0.440}
& \textbf{1.041} \\
\bottomrule
\end{tabular}
}
\end{table*}

\begin{figure}[htbp]
  \centering
  \includegraphics[width=0.5\textwidth]{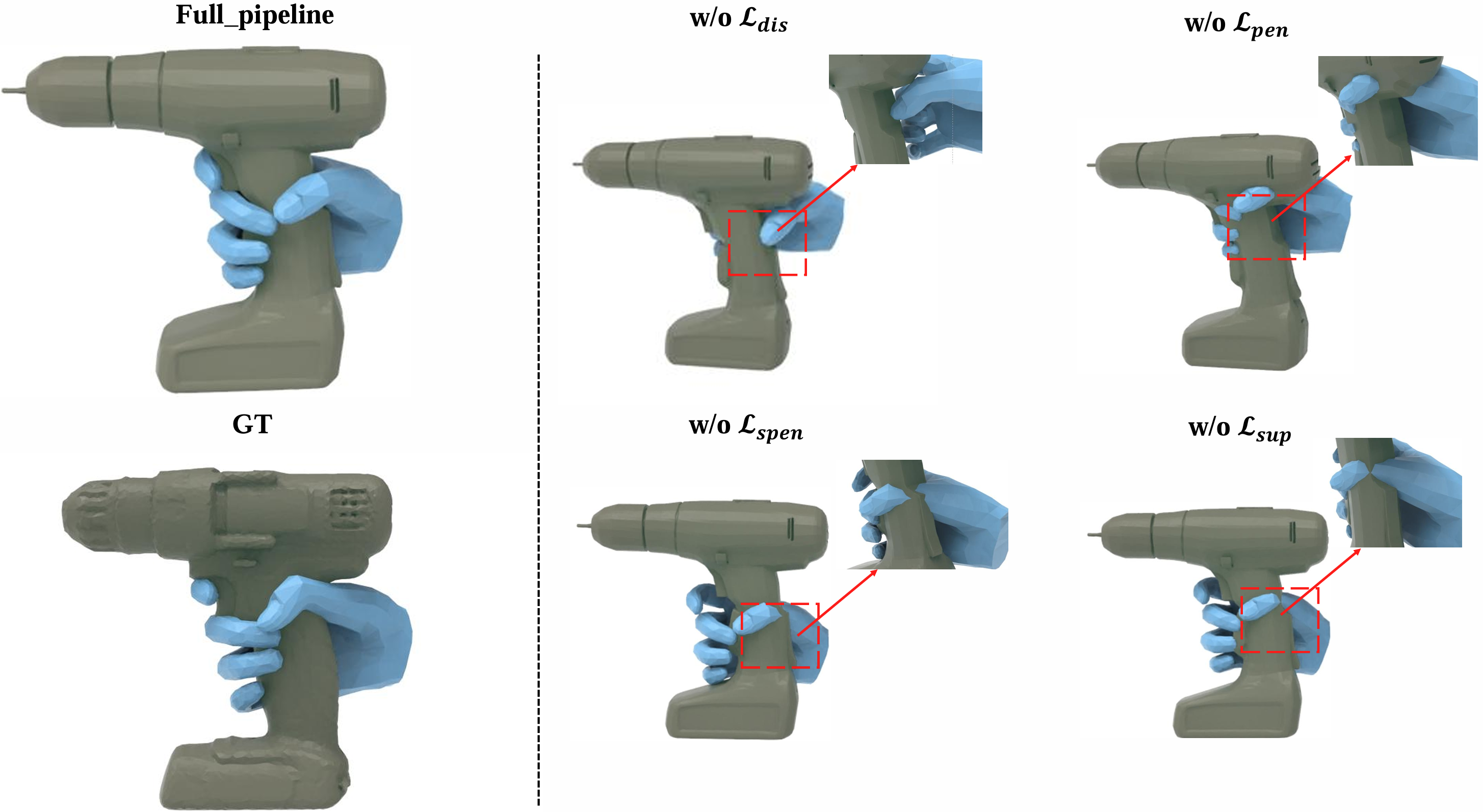} 
  \caption{Ablation study on hand-object contact visualization by removing individual components from $\mathcal{L}$.}

  \label{fig:abl}
\end{figure}

\begin{figure*}[htbp]
  \centering
  \includegraphics[width=1.0\textwidth]{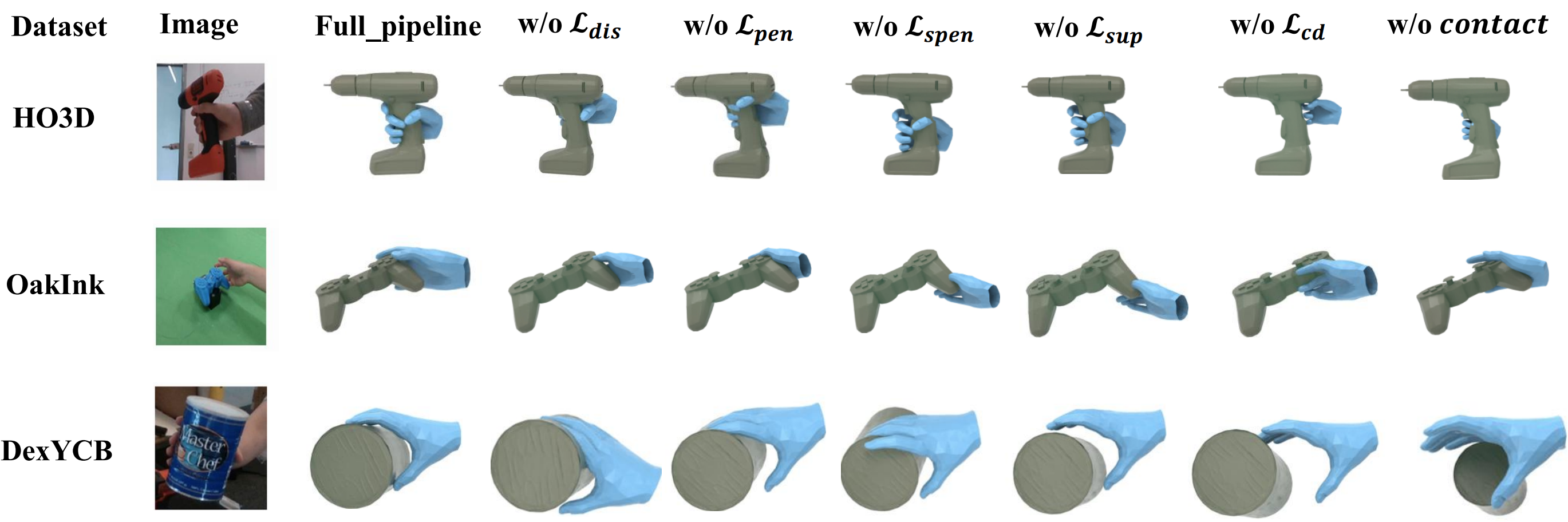}
  \caption{
  A visual analysis of the ablation study on the loss terms in $\mathcal{L}$, $\mathcal{L}_{\text{cd}}$ and contact optimization. In the experiments, each component is removed individually, and hand parameters are optimized afterward to analyze the specific impact of each term on the optimization performance.
  }
  \label{fig:ab}
\end{figure*}

To evaluate the role of the optimization losses, we conduct ablation studies by removing individual terms from the objective $\mathcal{L}$. The qualitative results in Fig.~\ref{fig:ab} show that removing key terms increases the spatial deviation between the hand and object. Fig.~\ref{fig:abl} further shows that removing individual loss components can introduce local penetration and unstable contact.

The quantitative results in Tab.~\ref{tab:ablation_study_metric1} are consistent with the visual observations. Although individual variants may improve one metric in isolation, the full pipeline achieves the best overall balance across SD, IPI, CR, and SIV. These results indicate that the proposed objective is necessary for balancing geometric alignment and physical plausibility.

The ablation of the OPA module further clarifies the role of projection-based alignment and contact optimization. As shown in Fig.~\ref{fig:ab}, removing $\mathcal{L}_{cd}$ leads to spatial misalignment between the hand and object, while removing contact optimization weakens effective hand-object contact. The quantitative results in Tab.~\ref{tab:ablation_study_metric2} show a consistent degradation in interaction metrics when either component is removed, confirming their importance for maintaining stable and physically plausible hand-object interactions.
\begin{table}[!t]
  \centering
  \footnotesize
  \caption{Quantitative sizes of generated objects for $\mathcal{L}_{cd}$ and contact optimization. Lower SD and SIV are better ($\downarrow$), while higher IPI and CR are better ($\uparrow$).}
  \label{tab:ablation_study_metric2}
  \resizebox{\linewidth}{!}{%
  \begin{tabular}{llcccc}
    \toprule
    \multicolumn{2}{c}{} & SD(cm) $\downarrow$ & IPI $\uparrow$ & CR(\%) $\uparrow$ & SIV($\times100cm^3$) $\downarrow$ \\
    \midrule
    Baseline & Full Pipeline   & \textbf{0.517} & \textbf{0.175} & 0.176 & 0.0092 \\
    \midrule
    \multirow{2}{*}{Module} & w/o $\mathcal{L}_{\text{cd}}$ & 0.873 & 0.130 & 0.131 & 0.0047 \\
    & w/o contact  & 9.027 & 0.157 & 0.174 & 0.1009 \\
    \bottomrule
  \end{tabular}
  }
\end{table}

\begin{table}[!t]
  \centering
  \caption{Ablation study results. We compare the full pipeline against variants where specific loss components are removed. Lower SD and SIV are better ($\downarrow$), while higher IPI and CR are better ($\uparrow$).}
  \label{tab:ablation_study_metric1}
  \resizebox{\linewidth}{!}{%
  \begin{tabular}{llcccc}
    \toprule
    \multicolumn{2}{c}{} & SD(cm) $\downarrow$ & IPI $\uparrow$ & CR(\%) $\uparrow$ & SIV($\times100cm^3$) $\downarrow$ \\
    \midrule
    Baseline & Full Pipeline   & \textbf{0.517} & \textbf{0.175} & 0.176 & 0.0092 \\
    \midrule
    \multirow{4}{*}{Loss Term} & w/o $\mathcal{L}_{\text{dis}}$  & 1.25 & 0.051 & 0.051 & \textbf{0.0001} \\
    & w/o $\mathcal{L}_{\text{pen}}$  & 2.45 & 0.155 & \textbf{0.178} & 0.1368 \\
    & w/o $\mathcal{L}_{\text{spen}}$ & 1.32 & 0.124 & 0.226 & 0.0115 \\
    & w/o $\mathcal{L}_{\text{sup}}$  & 1.08 & 0.156 & 0.156 & 0.0012 \\
    \bottomrule
  \end{tabular}
  }
\end{table}

\begin{figure}[htbp]
  \centering
  \includegraphics[width=0.45\textwidth]{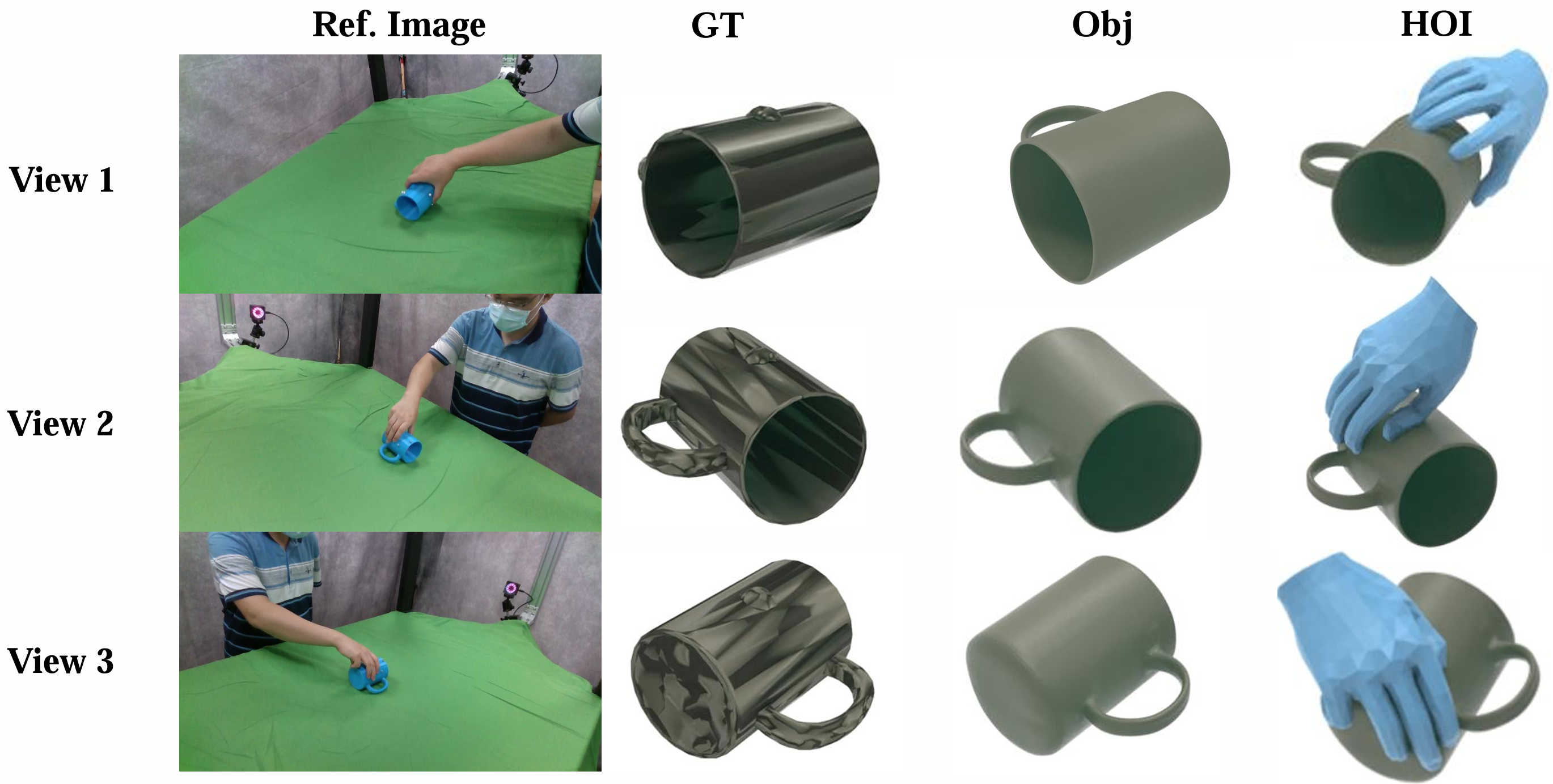} 
  \caption{
Qualitative visualization of viewpoint-robust 3D scale alignment. Although the input images exhibit substantial camera pose variations, our method preserves stable hand-object proportions and produces geometrically consistent HOI reconstructions across views.}

  \label{fig:scale}
\end{figure}

To analyze the robustness of our scale alignment strategy under different viewpoints, we conduct multi-view experiments on OakInk, where the same hand-object interaction is captured from three substantially different camera poses (View~1--3). As illustrated in Fig.~\ref{fig:scale}, each row corresponds to a distinct viewpoint, while the columns present the reference RGB image, the ground-truth object mesh (GT), the reconstructed object mesh (Obj), and the final HOI reconstruction (HOI). We quantify the relative 3D scale deviation as follows:

\begin{equation}
{
\text{Relative Proportion Error}
=
\frac{
\left|
\left(\frac{S_o}{S_h}\right)_{\text{Ours}}
-
\left(\frac{S_o}{S_h}\right)_{\text{GT}}
\right|
}{
\left(\frac{S_o}{S_h}\right)_{\text{GT}}
}
}
\end{equation}

where $S_o$ and $S_h$ denote the 3D spatial scales of the object and hand, respectively, computed as the diagonal length of their axis-aligned bounding boxes. Due to the inherent global scale ambiguity in monocular reconstruction, moderate deviations in absolute object scale are expected. Nevertheless, both the quantitative metrics reported in Table~\ref{tab:scale} and the qualitative visualization in Fig.~\ref{fig:scale} demonstrate stable reconstruction performance across viewpoints. Although the camera poses vary significantly, the reconstructed objects preserve consistent geometric proportions relative to the hand, and the resulting HOI scenes maintain coherent contact relationships. These results indicate that our 2D-guided scale alignment mechanism effectively stabilizes hand-object proportions under viewpoint changes, thereby ensuring functional plausibility in single-view HOI reconstruction.

\begin{table}[htbp]
\centering
\caption{
Quantitative comparison of hand-object relative proportion consistency 
and HOI reconstruction quality under different viewpoints.
}
\label{tab:scale}
\resizebox{\linewidth}{!}{
\begin{tabular}{c|c|c|c|c}
\hline
Viewpoint 
& Relative Proportion Error $\downarrow$ 
& F@5 $\uparrow$ 
& F@10 $\uparrow$ 
& CD $\downarrow$ \\
\hline
View 1 & 0.12 & 0.34 & 0.50 & 0.99 \\
View 2 & 0.09 & 0.37 & 0.54 & 0.93 \\
View 3 & 0.15 & 0.37 & 0.51 & 0.95 \\
\hline
Average & 0.12 & 0.36 & 0.52 & 0.96 \\
\hline
\end{tabular}
}
\end{table}

\begin{figure}[htbp]
  \centering
  \includegraphics[width=0.45\textwidth]{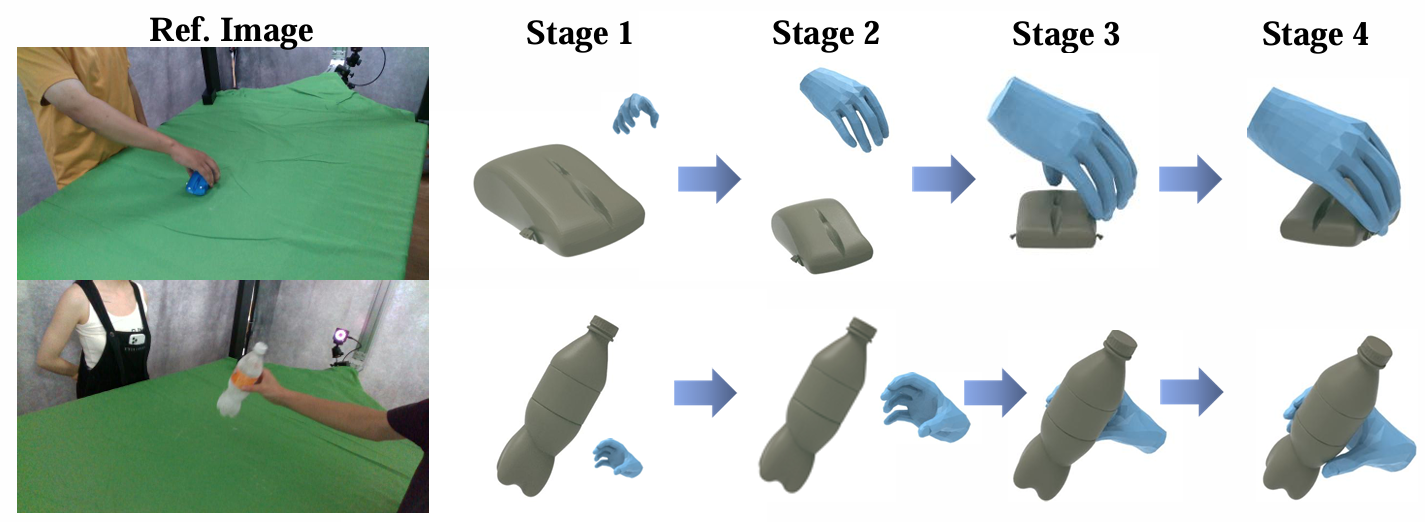} 
\caption{
Stage-wise qualitative ablation of our pipeline. 
Stage~1 performs independent hand and object mesh reconstruction. 
Stage~2 aligns the relative scale between hand and object. 
Stage~3 applies OPA to correct spatial alignment. 
Stage~4 further refines contact to produce physically plausible HOI.}

  \label{fig:stage}
\end{figure}

To further analyze the contribution of each module, we present a stage-wise qualitative ablation in Fig.~\ref{fig:stage}. 
Stage~1 reconstructs the hand and object meshes independently, leading to scale inconsistencies and spatial misalignment. 
Stage~2 aligns the relative scale between the hand and object, correcting global size discrepancies. 
Stage~3 incorporates the OPA, which refines spatial alignment through 2D projection consistency and improves geometric consistency. 
Stage~4 further refines the interaction via contact-aware optimization, removing interpenetration artifacts and enhancing physical plausibility. 
The progressive refinement from Stage~1 to Stage~4 demonstrates that scale alignment and OPA ensure geometric consistency, while the contact optimization module is essential for enforcing physically plausible hand-object interaction.

\section{Discussion}

\begin{figure}[htbp]
  \centering
  \includegraphics[width=0.45\textwidth]{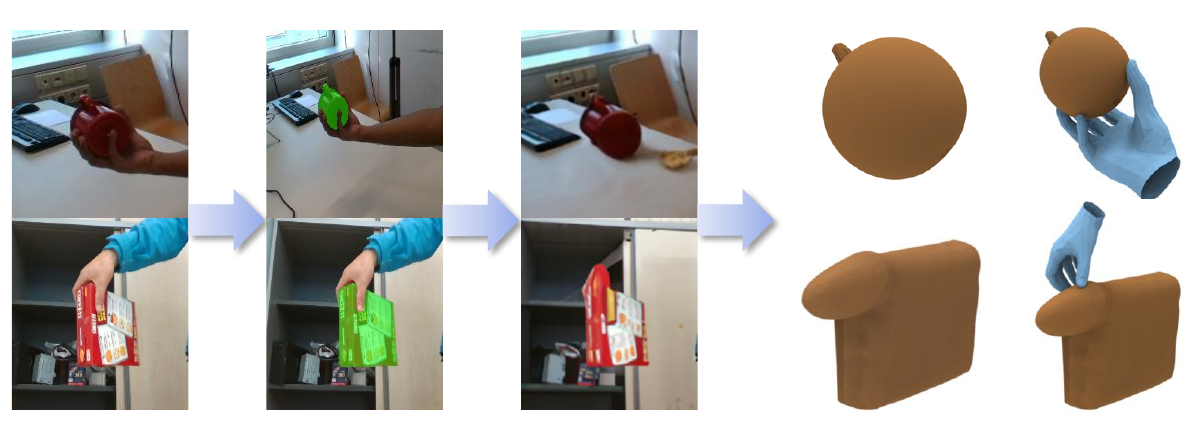} 
  \caption{
Failure case caused by severe hand--object occlusion.
Heavy occlusion introduces ambiguity in object geometry recovery, leading to inaccurate reconstruction and error propagation in the final 3D HOI result.
}
  \label{fig:fail1}
\end{figure}

Fig.~\ref{fig:fail1} presents a failure case caused by severe hand--object occlusion.
When a large portion of the object is occluded by the hand, the inpainting stage may recover an inaccurate object appearance.
Since the subsequent object reconstruction relies on the completed object image, such errors can propagate to the reconstructed object mesh and degrade the final 3D HOI result.
This failure mode reflects the geometric ambiguity of object completion under heavy occlusion.

\begin{figure}[htbp]
  \centering
  \includegraphics[width=0.45\textwidth]{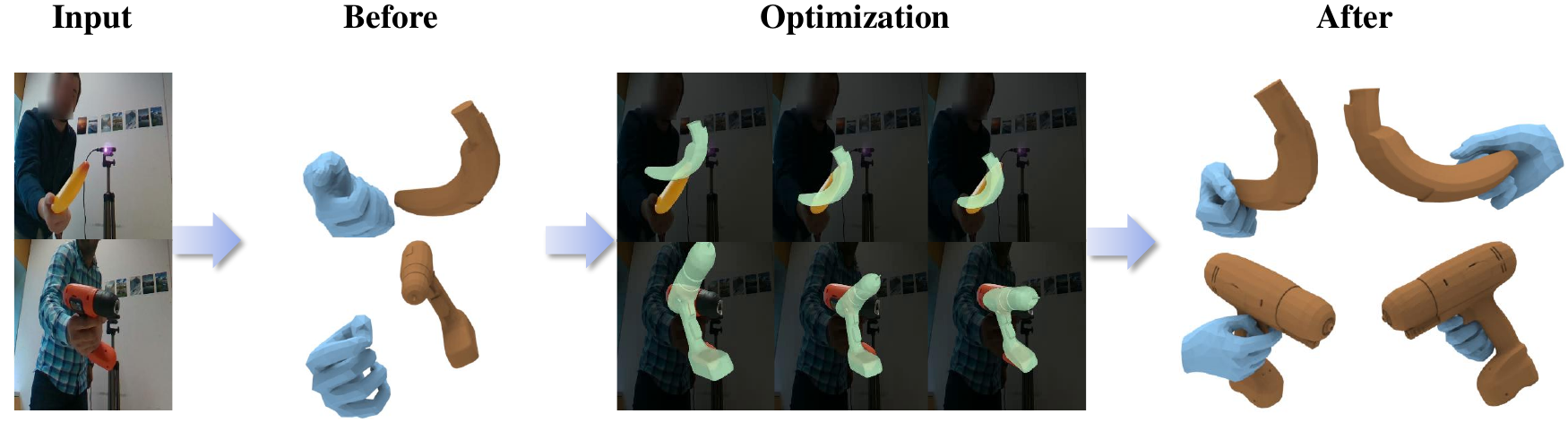} 
  \caption{
Failure case caused by inaccurate hand--object initialization.
Although the reconstructed object appears consistent with the 2D observation, the single-view setting does not fully constrain the 3D relative pose.
The resulting initialization error may lead OPA to a suboptimal alignment, which further limits the effectiveness of the subsequent contact-aware refinement.
}
  \label{fig:fail2}
\end{figure}

Fig.~\ref{fig:fail2} presents a second failure case caused by inaccurate hand--object initialization.
In FunHOI, the Object Pose Approximator (OPA) estimates the object pose mainly by enforcing consistency with the single-view 2D observation.
However, different 3D hand--object configurations may produce similar 2D projections.
Consequently, the hand and object can appear aligned in the image plane while still being misaligned in 3D space.
Without multi-view observations or additional 3D constraints, OPA may converge to a suboptimal initialization, which limits the effectiveness of subsequent contact-aware refinement.

These cases indicate that the main limitations of the current framework arise from severe occlusion, geometric ambiguity in single-view reconstruction, and initialization sensitivity.
Future work may incorporate more robust object completion, multi-view constraints, or stronger 3D interaction priors to improve robustness under these challenging conditions.

Furthermore, our approach focuses on generating static 3D functional HOI models and has not yet been extended to continuous dynamic 3D functional HOI sequences.

Extending functional text guidance from static HOI configurations to temporally coherent 3D interaction sequences remains an important direction for future work, particularly for applications in embodied intelligence, simulation, and controllable 3D content generation.

\section{Conclusion}

We presented FunHOI, a two-stage framework for functional-intent-conditioned 3D hand--object interaction synthesis without requiring 3D HOI annotations. 
FunHOI combines a Functional Grasp Generator (FGG), which produces semantically aligned 2D HOI guidance and initial hand--object reconstructions, with a Functional Grasp Refiner (FGR), which performs scale-aware alignment, object pose approximation, and contact-aware refinement. 
Rather than learning a direct text-to-3D generative prior, FunHOI uses functional text guidance as an intermediate semantic condition for reconstructing and composing physically plausible 3D hand--object interactions.
By representing functional intent as action--object--purpose prompts, the framework supports task-oriented grasp configurations that depend on the intended object use rather than only on object category or positional commands. 
Extensive experiments show that FunHOI achieves competitive reconstruction quality, improved hand--object contact plausibility, and stronger functional-intent alignment compared with representative reconstruction-based and text-guided baselines. 
We also identify current limitations under severe occlusion, single-view geometric ambiguity, and initialization-sensitive cases, which suggest future directions including more robust object completion, multi-view constraints, dynamic interaction modeling, and stronger 3D interaction priors.

%

\vfill
\begin{IEEEbiography}[{\includegraphics[width=1in,height=1.25in,clip,keepaspectratio]{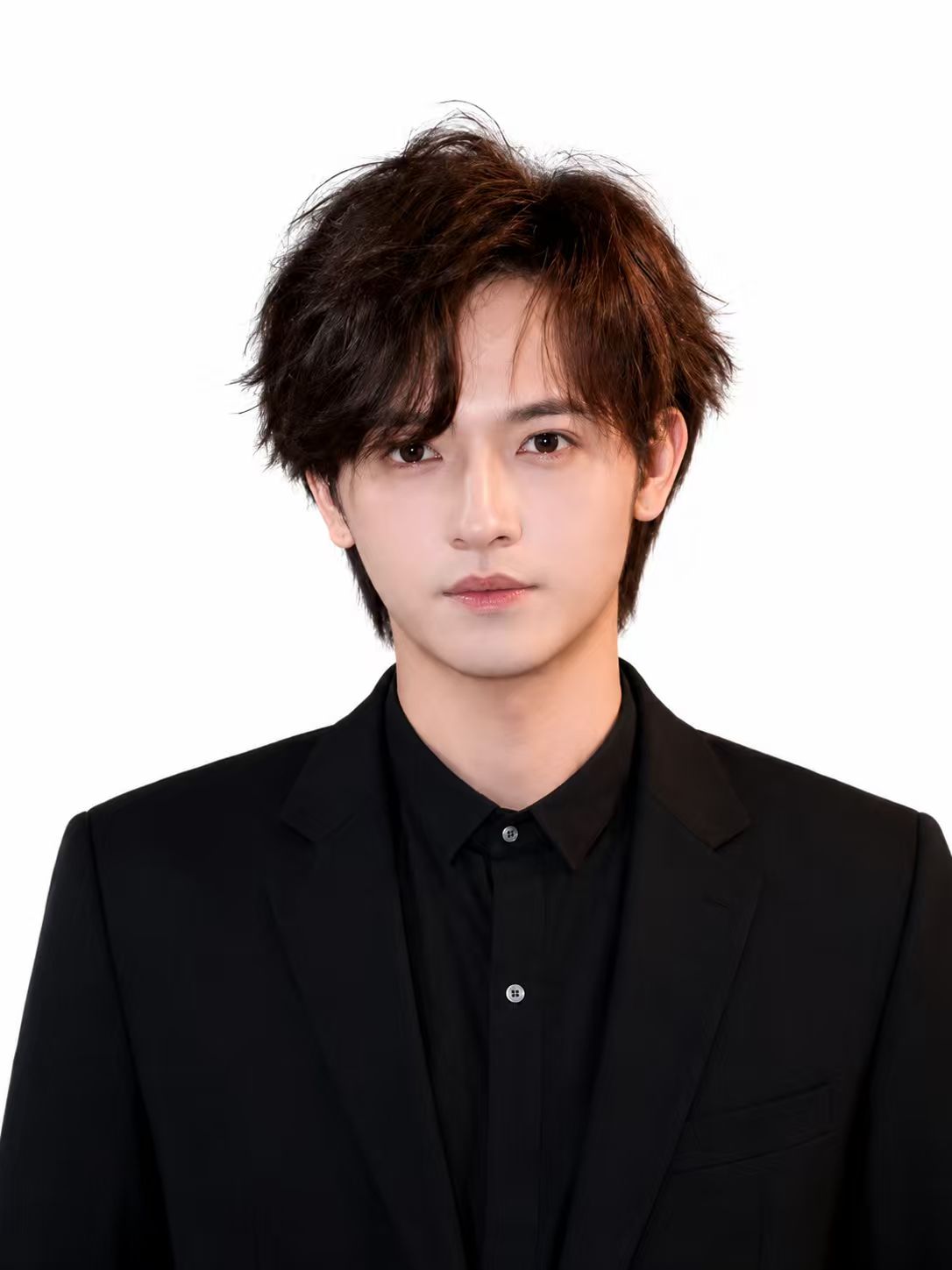}}]{Yongqi Tian}
Yongqi Tian received the B.S. and M.S. degrees from Beijing Institute of Technology, Beijing, China, in 2019 and 2022, respectively. He is currently pursuing the Ph.D. degree with the School of Artificial Intelligence, Xi'an Jiaotong University, Xi'an, China.
\end{IEEEbiography}

\begin{IEEEbiography}[{\includegraphics[width=1in,height=1.25in,clip,keepaspectratio]{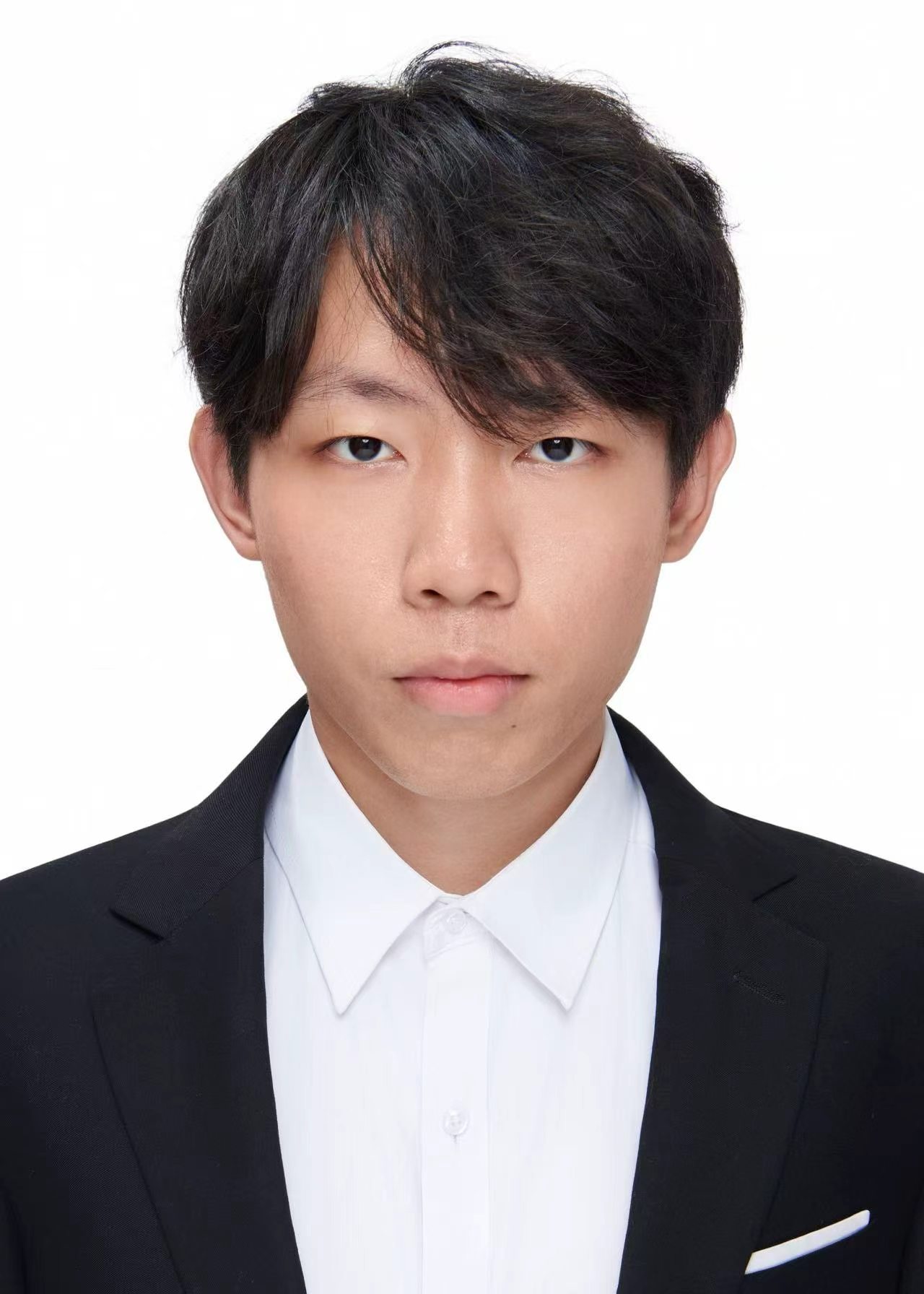}}]{Xueyu Sun}
Xueyu Sun received the B.S. degree from Jilin University, Changchun, China, in 2023. She is currently pursuing the M.S. degree with Xi'an Jiaotong University, Xi'an, China. 
\end{IEEEbiography}

\begin{IEEEbiography}[{\includegraphics[width=1in,height=1.25in,clip,keepaspectratio]{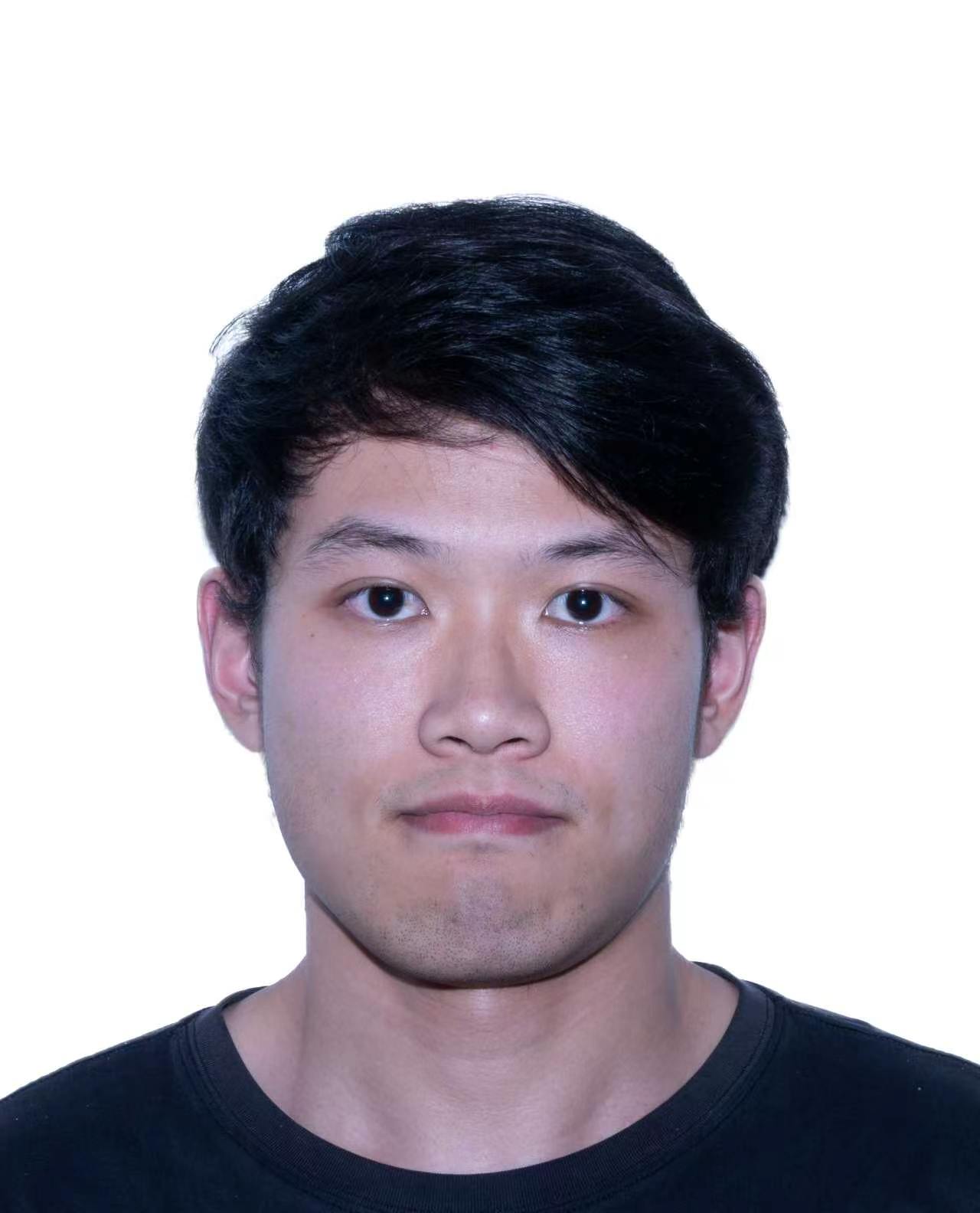}}]{Haoyuan He}
Haoyuan He received the B.S. degree from Xi'an Jiaotong University, Xi'an, China, in 2024. He is currently pursuing the Ph.D. degree with Xi'an Jiaotong University, Xi'an, China. 
\end{IEEEbiography}

\begin{IEEEbiography}[{\includegraphics[width=1in,height=1.25in,clip,keepaspectratio]{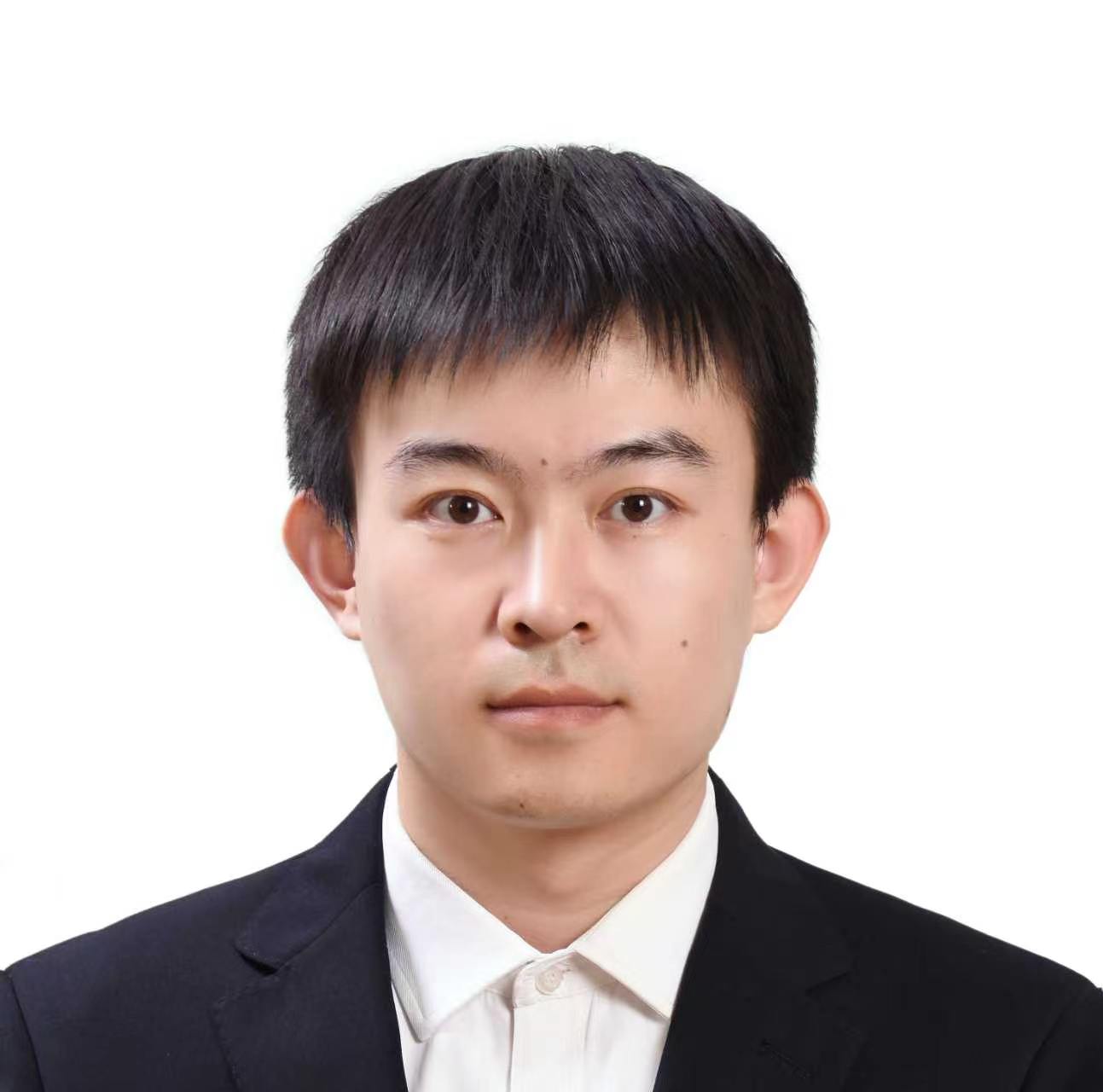}}]{Jianlei Wang}
Jianlei Wang received the B.S. degree from Wuhan University of Technology, Wuhan, China, in 2022. He is currently pursuing the Ph.D. degree with Xi'an Jiaotong University, Xi'an, China.
\end{IEEEbiography}

\begin{IEEEbiography}[{\includegraphics[width=1in,height=1.25in,clip,keepaspectratio]{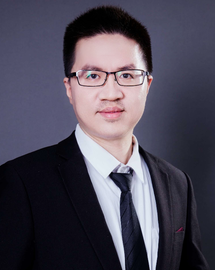}}]{Caigui Jiang}
Caigui Jiang is a Professor with the Institute of Artificial Intelligence and Robotics, Xi'an Jiaotong University (XJTU), Xi'an, China. He received the B.S. and M.S. degrees from XJTU in 2008 and 2011, respectively, and the Ph.D. degree from King Abdullah University of Science and Technology (KAUST), Saudi Arabia, in 2016. Prior to joining XJTU, he was a Research Scientist and Postdoctoral Researcher with the Visual Computing Center (VCC), KAUST, Saudi Arabia; the International Computer Science Institute (ICSI), University of California, Berkeley, USA; and the Max Planck Institute for Informatics, Germany. His research interests include computer graphics, geometric modeling, intelligent vehicles, and robotics.
\end{IEEEbiography}

\end{document}